\documentclass[a4paper,fleqn]{cas-dc}

\usepackage{soul} 
\usepackage{color, xcolor}

\usepackage[authoryear]{natbib}

\usepackage{subfigure}
\usepackage{tabularx}
\usepackage{tikz}
\usepackage{pgfplots}
\usepackage[edges]{forest}
\usepackage{adjustbox}

\usepackage{tabularx}

\usepackage{bm}

\usepackage{pifont}
\newcommand{\cmark}{\ding{51}} % ✓
\newcommand{\xmark}{\ding{55}} % ✗

\begin{document}
\shorttitle{Large Language Models for Robotics: A Survey} 
\shortauthors{F. Zeng \textit{et al.}}

\author[1]{Fanlong Zeng}
\ead{flzeng1@gmail.com}

\author[1]{Wensheng Gan}
\ead{wsgan001@gmail.com}
\cormark[1]

\author[2, 3]{Zezheng Huai}
\ead{huaizz24@mails.jlu.edu.cn}

\author[4]{Lichao Sun}
\ead{lis221@lehigh.edu}

\author[2]{Hechang Chen}
\ead{chenhc@jlu.edu.cn}

\author[1]{Yongheng Wang}
\ead{yonghengwwang@gmail.com}

\author[1]{Ning Liu}
\ead{tliuning@jnu.edu.cn}

\author[5]{Philip S. Yu}
\ead{psyu@uic.edu}

\address[1]{School of Intelligent Systems Science and Engineering, Jinan University, Zhuhai 519070, China}
\address[2]{School of Artificial Intelligence, Jilin University, Changchun, China}
\address[3]{Shanghai Innovation Institute, Shanghai, China}
\address[4]{Lehigh University, Bethlehem, USA}
\address[5]{University of Illinois Chicago, Chicago, USA}

\cortext[cor1]{Corresponding author}

\title [mode = title]{Large Language Models for Robotics: A survey}

\begin{abstract}
    The human ability to learn, generalize, and control complex manipulation tasks through multi-modality feedback suggests a unique capability, which we refer to as dexterity intelligence. Understanding and assessing this intelligence is a complex task. Amidst the swift progress and extensive proliferation of large language models (LLMs), their applications in the field of robotics have garnered increasing attention. LLMs possess the ability to process and generate natural language, facilitating efficient interaction and collaboration with robots. Researchers and engineers in the field of robotics have recognized the immense potential of LLMs in enhancing robot intelligence, human-robot interaction, and autonomy. Therefore, this comprehensive review aims to summarize the applications of LLMs in robotics, delving into their impact and contributions to key areas such as robot control, perception, decision-making, and planning. This survey first provides an overview of the background and development of LLMs for robotics, followed by a discussion of their benefits and recent advancements in LLM-based robotic models. It then explores various techniques, employed in perception, decision-making, control, and interaction, as well as cross-module coordination in practical tasks. Finally, we review current applications of LLMs in robotics and outline potential challenges they may face in the near future. Embodied intelligence represents the future of intelligent systems, and LLM-based robotics is one of the most promising yet challenging paths toward achieving it.
\end{abstract}


\maketitle

\begin{figure*}[h]
    \centering
    \includegraphics[scale=0.40]{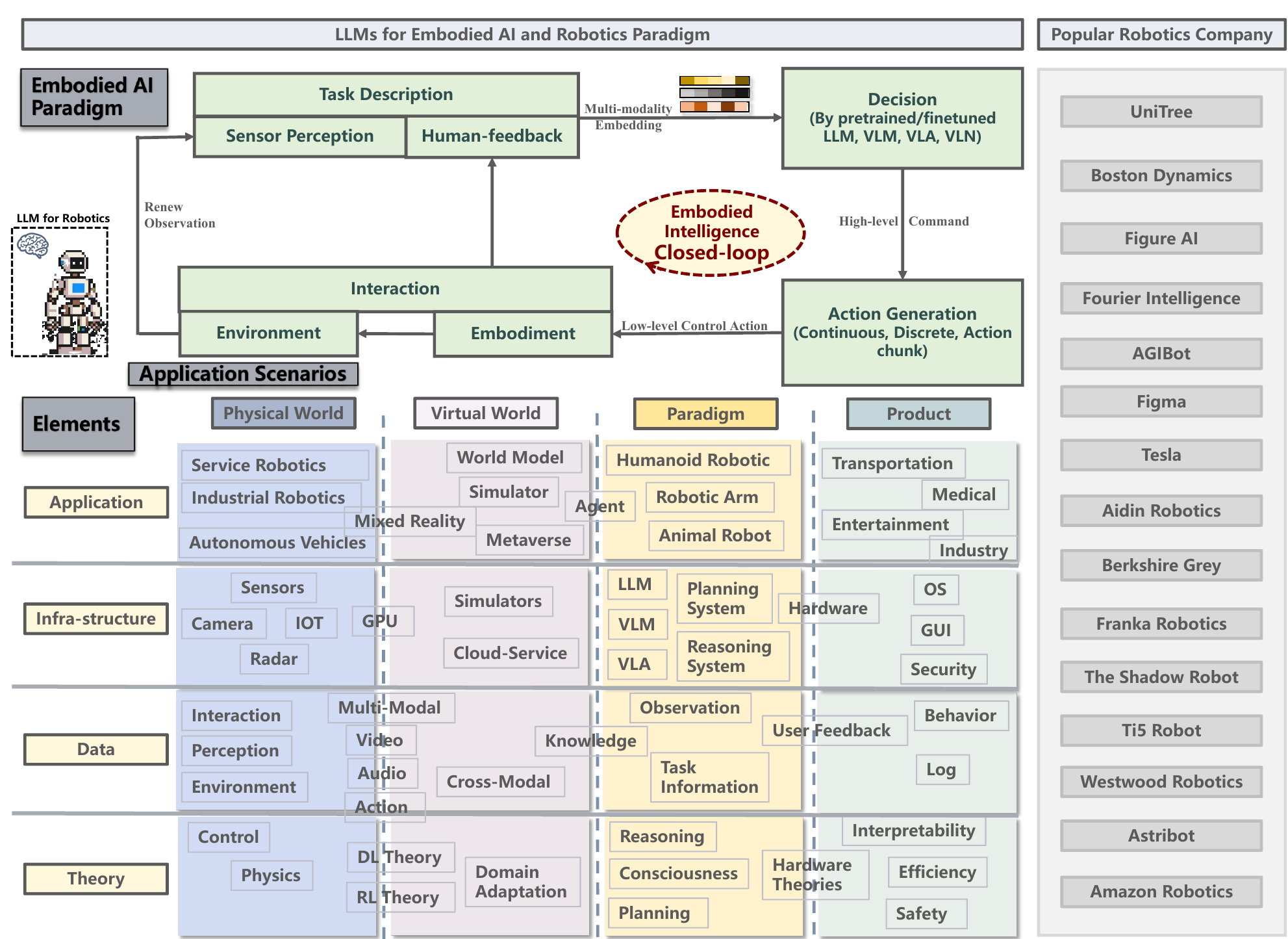}
    \caption{Overview of embodied robotics systems that perceive and interact with diverse environments. Through physical embodiment, these systems integrate perception, decision-making, and task execution, enabling a shift from purely virtual cognition to direct physical interaction. Embodied robotics represents a critical transformation in AI, bridging the gap between abstract reasoning and real-world deployment. In this work, we present a comprehensive overview of embodied robotics.}
    \label{fig: overview}
\end{figure*}

\section{Introduction}\label{sec:Introduction}

Humans possess exceptional proficiency in executing intricate and dexterous manipulation skills by integrating tactile, visual, and other sensory inputs. Research in the field of robotics aspires to imbue robots with comparable manipulation intelligence. Although recent advancements in robotics and machine learning have yielded promising results in visual mitigation and exploration learning for robot manipulation, there remains much to be accomplished in this area. Large language models (LLMs), such as DeepSeek \citep{deepseekai2024deepseekv3, wu2024deepseekvl2}, GPT-5\citep{GPT5}, Llama4 \citep{meta2025llama} have emerged as significant research achievements in the field of artificial intelligence (AI) in recent years. Through deep learning techniques \citep{lecun2015deep}, LLMs can be trained on massive text corpora, enabling them to generate high-quality natural language text. This development has sparked new thinking in natural language processing and dialogue systems. At the same time, the rapid advancement of robotics technology \citep{karoly2020deep,dorigo2021swarm} has created a demand for more intelligent and natural human-machine interaction. Combining LLMs with robots can provide robots with stronger natural language understanding and generation capabilities, enabling more intelligent and human-like conversations and interactions. We have presented the overview of robotics combined with LLM in Fig. \ref{fig: overview}.

Applying LLMs to the field of robotics has important research significance and practical value. Firstly, LLMs can significantly enhance a robot's natural language understanding and generation capabilities. Traditional robot dialogue systems often rely on manually crafted rules and templates, making it difficult for them to handle complex natural language inputs. LLMs, on the other hand, can better understand and generate natural language by learning from massive text corpora, enabling robots to have more intelligent and natural conversation abilities. Secondly, LLMs can provide more diverse conversation content and personalized interaction experiences. Through interaction with LLMs, robots can generate varied responses and personalize interactions based on user preferences and needs. This helps improve user satisfaction and interactions. In addition, the combination of LLMs and robots contributes to the advancement of AI and robotics technology, laying the foundation for future intelligent robots (or called smart robots).

%Current Research Status:
Currently, many research teams and companies have begun exploring the application of LLMs in the field of robotics. Some research focuses on using LLMs for natural language understanding in robots. By using pre-trained language models \citep{zeng2023distributed}, robots can better understand user intentions and needs \citep{ driess2023palm, shah2023lm}. Other research focuses on using LLMs for natural language generation in robots. Robots can generate fluent and coherent natural language responses through interaction with language models. Furthermore, some research explores how to combine LLMs with other technologies, such as knowledge graphs and sentiment analysis, to further enhance robot dialogue capabilities and user experiences \citep{qi2024safety}. From multiple perspectives, LLMs-based robotics is one of the most promising paths to achieve embodied intelligence in the future.

Although the combination of LLMs and robots has many potential advantages, it also faces challenges and issues \citep{helberger2023chatgpt, mccallum2023chatgpt}. Firstly, training and deploying LLMs require substantial computing resources and large-scale data, which can be challenging for resource-limited robot platforms \citep{overviewofagents}. Secondly, LLMs may generate inaccurate, unreasonable, or even harmful content when generating natural language text. Therefore, effective filtering and control mechanisms are essential to ensure that the content generated by robots adheres to ethical and legal requirements \citep{mccallum2023chatgpt}. Additionally, robot dialogue systems need to address challenges such as multi-turn dialogues, context understanding, and dialogue consistency to provide more coherent and human-like interactions. Furthermore, the shape of robots has not been standardized across the industry. The question remains whether robots should adopt a humanoid form or take on a different shape \citep{hwang2013effects}. In other words, what form of robot is best suited for our needs? The impact of embodied intelligence on our society cannot be overstated. Will robots eventually replace human labor? How should we respond to this seismic shift in the future? Moreover, if robots were to gain consciousness, should we still view them as tools? How should humans define a conscious robot?

In conclusion, the applications of LLMs in robotics hold tremendous potential. They provide new paradigms and methods for robot control, path planning, and intelligence. Through more intuitive and natural human-machine interaction, language-based path planning, and intelligent semantic understanding, LLMs not only enhance the performance and efficiency of robots but also improve the experience and interaction modes of human-robot interaction. This comprehensive review aims to summarize the applications of LLMs in robotics, examining their impact and contributions to key areas such as robot control, perception, decision-making, and path planning. To summarize, there are four key contributions in this paper, as follows:

\begin{itemize}	
    \item This survey discusses the latest advancements in LLMs and their significant impact on the field of robotics. It highlights the benefits of LLMs for robotic systems and examines the emergence of new robot models equipped with LLMs in recent years.

    \item This survey discussed the current state of robot technology, focusing on advancements in perception, decision making, control, and interaction combined with LLMs. Specifically, we highlighted the critical role of LLMs in decision-making modules, which have enabled robots to make more informed and effective decisions in various applications.

    \item This survey explored potential applications of current robots equipped with LLMs in the near future. 
    
    \item This survey discussed several potential challenges that robots may face when integrated with LLMs, as well as the possible impacts of future developments in this field on human society. 
\end{itemize}

% \textbf{Organization}
The rest of this article is organized as follows. In Section \ref{sec:overview}, a concise review of the development history of LLMs is provided. Additionally, this section also introduces a selection of prominent LLM companies and their flagship products. Section \ref{sec: LLM for Robotics} introduces representative robotics based on LLM, including agents and embodied robotic companies with its corresponding flagship products. In Section \ref{sec:technologies}, core modules of robotics are introduced. Additionally, the cross coordination of these modules is illustrated in Section \ref{sec: cross-module coordination}. The applications of robotics based on LLM are introduced in Section \ref{sec: applications}. Moreover, this paper highlights the challenges in Section \ref{sec: challenge} and presents several promising directions of LLMs for robotics in Section \ref{sec:directions}. Finally, we conclude this paper in Section \ref{sec:conclusion}.

\section{Language Model Overview}
\label{sec:overview}

Amidst the swift progress and extensive proliferation of LLMs, the model of robotics based on LLMs has emerged. LLM serves as a robotics brain, like in Figure \ref{fig: robotics combined with LLMs}, making it more intelligent. In this part, We first provide an overview of LLM, and delve into the history of LLM's development, followed by an introduction to the derivatives widely used in robotics.

\begin{table}[h]
    \caption{Abbreviation with its corresponding description}
    \small
    \label{tab:abbr with des}
    \begin{tabular}{l|l}
    \hline
    \textbf{Abbreviation} & \multicolumn{1}{c}{\textbf{Description}} \\ \hline
    B & Billion \\
    AI  & Artificial Intelligence      \\
    GPT & Generative Pre-trained Transformer \\
    VNM & Vision-Navigation Model    \\
    VLM & Vision-Language Model     \\
    VLN & Vision-and-Language Navigation model   \\
    VLA & Vision-Language-Action model  \\ 
    MAS & Multi-Agent Systems \\ 
    LLMs & Large-scale Language Models      \\
    MLLMs & Multimodal Large Language Models\\

    \hline
    \end{tabular}
\end{table}

\subsection{What is LLM}

A language model is a computational model that utilizes statistical methods to analyze and predict the probability of word sequences in a given language. It is designed to capture the patterns, grammar, and semantic meaning of natural language \citep{DBLP:journals/corr/abs-2303-08774}. While LLM is a type of artificial intelligence designed to comprehend and generate human language, marking a significant milestone in the pursuit of Artificial General Intelligence (AGI) \citep{zhao2023survey}. In contrast to traditional models, LLMs boast exceptional capabilities, including: an exceptionally broad knowledge base, acquired through extensive pre-training on vast amounts of text corpora, which enables LLMs to possess a deep understanding of human language; outstanding performance in a wide range of text-based tasks, making them an effective tool for serving as capable assistants; a certain certain level of reasoning ability, enabling them to solve simple mathematical problems \citep{zhao2023survey}. However, its logical reasoning capabilities still have significant room for improvement and remain an area that warrants further exploration. Recent advancements have also led to the development of LLMs that can acquire the ability to perform task planning and decompose complex problems \citep{xue2024decompose, peng2024chain}. Furthermore, LLMs can leverage external tools, such as calculators. Due to their exceptional capabilities, LLMs can effectively cater to the requirements of various downstream tasks, such as text and code generation, as well as completion tasks. Moreover, LLMs can be easily adapted to specific domains through simple fine-tuning, enabling them to excel in specialized tasks \citep{hu2022lora}. Notably, LLMs can achieve few-shot learning in specific domains with minimal domain-specific data, and can also leverage dialogue history to facilitate in-context learning. Furthermore, they can even demonstrate zero-shot learning capabilities, allowing them to generalize to new tasks without requiring any additional training data \citep{GPT5, meta2025llama}. Owing to the remarkable capabilities of LLMs, researchers have started to explore the possibility of leveraging LLMs as the brain of embodied intelligent systems, trying to create the embodied intelligent \citep{yang2025agentic, vats2024recoverychaining}.

\subsection{Development of LLMs}
In this section, we provide a brief overview of the development of LLMs. Additionally, we introduce derivative models of LLM and provide the historical development of these models.

\begin{figure}
    \centering
    \includegraphics[scale=0.52]{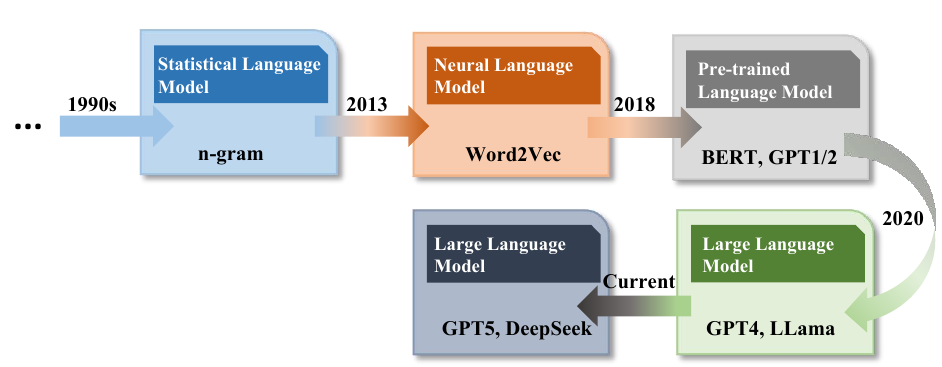}  
     \caption{The evolution and representative models (or methods) of LLM. Noticed that the time period of each distinct may not be accurate.}
    \label{fig: evoluation of LLM}
\end{figure}

\subsubsection{Evolution of Language Models}
 The development of LLMs can be roughly divided into four periods: (1) the Statistical Language Model (SLM) period, (2) the Neural Language Model (NLM) period, (3) the Pre-trained Language Model (PLM) period, and (4) the LLM period. The evoluation of LLM is summarized in Fig. \ref{fig: evoluation of LLM}. We present a detailed description of these four models: (1) SLM is a fundamental model in NLP. From a probabilistic and statistical perspective, it is a mathematical model that captures the contextual characteristics of natural language. SLM predicts the target word based on a fixed-length prefix. A statistical language model with a fixed context length is commonly referred to as an n-gram language model. However, SLMs suffer from the curse of dimensionality, which prevents them from accurately modeling complex high-order semantic relationships \citep{brown1992class, kneser1995improved}. (2) NLMs utilize neural networks to model word sequences and tackle the issue of the curse of dimensionality. During the era of NLMs, word embedding, a novel approach for modeling word sequences, was proposed. Word Embedding refers to the process of mapping a high-dimensional space of words into a lower-dimensional continuous vector space, where each word or phrase is represented as a dense vector in the real number field. This language modeling method, which leverages implicit semantic feature representation, offers a relatively universal solution for a wide range of NLP tasks \citep{bengio2003neural, mikolov2013efficient}. (3) PLMs are obtained by training on large-scale textual data, enabling them to understand and generate human-like natural language. One of the key advantages of PLMs is their excellent zero-shot capability, as well as their ability to perform well on specific downstream tasks after fine-tuning. A traditional PLM typically consists of two primary components: an encoder and a decoder. However, with the advent of the Transformer architecture \citep{vaswani2017attention}, PLMs have evolved into two distinct paradigms: encoder-only and decoder-only models, which marked the beginning of the era of LLMs \citep{devlin2018bert, radford2019language}. (4) LLM is a type of PLM characterized by an enormous number of parameters. Researchers have discovered that scaling up model parameters or data often yields improved performance on downstream tasks, a phenomenon that has been dubbed the ``Scaling Law". It's impressive that LLMs possess few-shot capabilities, leveraging in-context learning to effectively tackle downstream tasks. The capacity that distinguishes large models from their smaller counterparts is often referred to as ``Emergent Abilities". Currently, the majority of LLMs employ a decoder-only architecture \citep{brown2020language, kaplan2020scaling}.

\subsubsection{Derivative Models of LLM}
\begin{figure*}[t]
    \centering
    \includegraphics[scale=0.5]{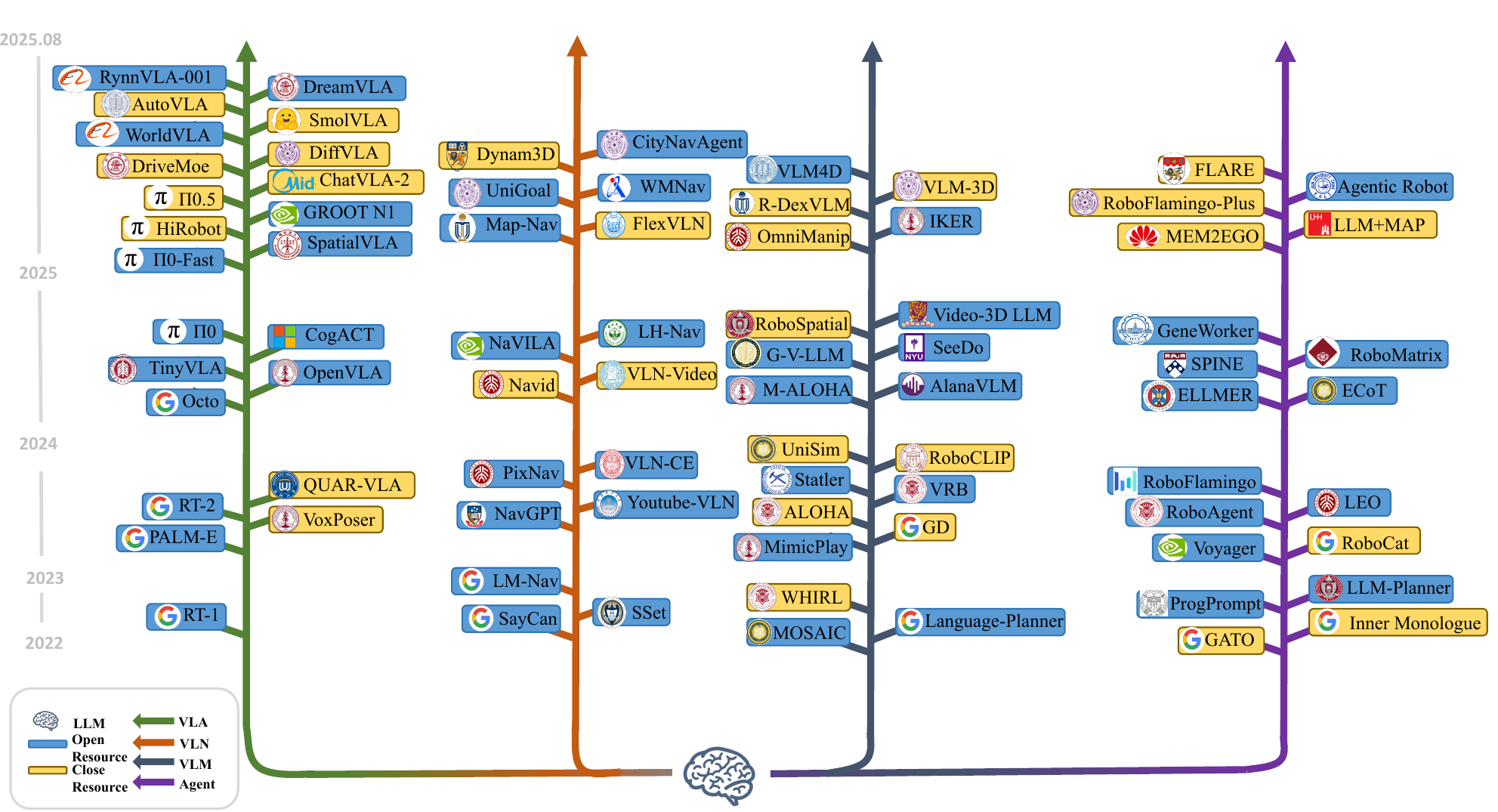}  
     \caption{The historical development of the Vision-Language-Action (VLA) model, Vision-Language-Navigation (VLN) model, Vision-Language-Model (VLM), and Agent. }
    \label{fig: historical development}
\end{figure*}

In this part, we introduce some derivative models of LLM. The derivative models can be categorized as (i) Vision-Language-Model (VLM) \citep{driess2023palm, zhou2025physvlm},  (ii) the Vision-Language-Action (VLA) model \citep{black2024pi_0, pertsch2025fast}, (iii) the Vision-Language-Navigation (VLN) model \citep{mirowski2018learning, ji2025dynavlm}, and (iv) Agent \citep{liu2025coherent, mandi2024roco}. The models are both derived from LLM, but the details of implementation and application fields are different. We have summarized a historical development of these derivatives in Fig. \ref{fig: historical development}.
Here, we introduce these different models: 
(i) The VLM represents a multimodal model capable of jointly processing and semantically aligning visual and linguistic information \citep{dosovitskiy2020image, chen2022pali}. These models exhibit three core capabilities: cross-modal semantic association,  visual content parsing with natural language generation, and multimodal reasoning for complex query resolution. VLMs have demonstrated significant utility across diverse applications, including automated image captioning, visual question answering (VQA), and medical image diagnosis.
(ii) The VLA model represents a multimodal model designed to integrate visual perception, linguistic comprehension, and physical actuation for robotics. Processing visual information and natural language instructions as inputs, VLA generates executable action sequences as outputs \citep{brohan2022rt, kim24openvla}. This architecture enables robots to perceive environmental stimuli,  interpret task requirements, and execute appropriate physical responses --- effectively bridging the gap between AI cognition and real-world interaction. VLA is potentially a critical advancement toward achieving embodied intelligence.
(iii) The VLN is a multimodal model designed to empower robotics and Unmanned Aerial Vehicles (UAVs) with the capability to interpret natural language commands and execute precise navigation within both physical and virtual 3D environments \citep{ling2025endowing, zhu2017target}. The core of the VLN is visual environment perception,  linguistic instruction comprehension, and dynamic path planning and execution. This technology requires the integration of computer vision (CV), natural language processing (NLP), spatial reasoning, and real-time decision-making capabilities. While VLN represents a crucial research direction in embodied AI, its primary applications are currently focused on autonomous navigation systems. Future developments may involve synergistic integration with VLA frameworks to achieve more comprehensive embodied intelligence.
(iv) The Agent is an autonomous entity capable of environmental perception, independent decision-making, and goal-directed action execution \citep{vats2024recoverychaining, yang2025agentic}. These agents may exist as software systems or embodied hardware systems (e.g., robotics platforms, virtual assistants). Their core functionality involves dynamic environment interaction and task completion through algorithmic processing, exhibiting three defining characteristics:  autonomy, adaptability, and collaborative capacity. Agents are mainly applied in highly repetitive and low-creativity work \citep{yuan2025remac}.

In summary, VLM, VLA, VLN, and Agent are critical derivative models of LLM that represent the future direction of embodied AI. We have also systematically compiled the historical development trajectories of both models for each derivative model in Table~\ref{tab:model_adaptation VLA} for VLA, Table~\ref{tab:model_adaptation VLN} for VLN, Table~\ref{tab:model_adaptation VLM} for VLM, and Table~\ref{tab:model_adaptation Agent} for Agent, respectively.

\begin{table*}[!ht]
\centering
\caption{VLA historian}
\label{tab:model_adaptation VLA}
\tiny  % 最小字体
\setlength{\tabcolsep}{0.5pt}  % 减小列间距
\renewcommand\arraystretch{0.8}
%\resizebox{\textwidth}{!}{
\begin{tabular}{@{}lccccccccccc@{}}
\toprule
& \multicolumn{8}{c}{Adaptation} & & \multicolumn{2}{c}{Evaluation} \\ % 合并第2到第9列
\cmidrule(lr){2-10}

\cmidrule(lr){11-12}

Model & \begin{tabular}[c]{@{}c@{}}Release \\ Time\end{tabular} & \begin{tabular}[c]{@{}c@{}}Size \\ (B)\end{tabular} & \begin{tabular}[c]{@{}c@{}}Base \\ Model\end{tabular} & \begin{tabular}[c]{@{}c@{}}IT \end{tabular} & \begin{tabular}[c]{@{}c@{}}RLHF \end{tabular} &  \begin{tabular}[c]{@{}c@{}}Pre-train \\ Data\end{tabular} & \begin{tabular}[c]{@{}c@{}}Latest Data \\ Timestamp\end{tabular} & \begin{tabular}[c]{@{}c@{}}Hardware \\ (GPUs / TPUs)\end{tabular} & \begin{tabular}[c]{@{}c@{}}Training \\ Time\end{tabular} & ICL   & CoT \\ \\
\midrule
RT-1 \citep{brohan2022rt} & DEC-2022 & 0.035 & EfficientNet-B3 & \xmark & \xmark  & 340k trajectories & 2021-09 & - & -& \xmark & \xmark \\
PALM-E \citep{driess2023palm} & MAR-2023 & 12/84/562 & PALM & \cmark & \xmark & - & 2023-02 & - &- & \cmark & \cmark \\
VoxPoser \citep{huang2023voxposer} &JUL-2023&-&GPT-4&\xmark&\xmark&-&2023-06&0&0&\cmark&\cmark\\
RT-2 \citep{brohan2023rt} &JUL-2023& 55 &  PaLI-X/PaLM-E & \cmark&\cmark&-&2023-06&-&-&\cmark&\cmark\\
QUAR-VLA \citep{ding2024quar} & DEC-2023 & 0.03/8 & Fuyu-8B/EfficientNet-B3 & \cmark & \xmark & 351K episodes & 2023-10 & - & - & \cmark & \cmark \\
Octo \citep{team2024octo} & MAY-2024 & 0.027/0.093 & Transformer & \cmark & \xmark & 800K trajectories & 2024-03 & - & - & \cmark & \cmark \\
OpenVLA \citep{kim24openvla} & JUN-2024 & 7 & Prismatic-7B/Llama 2 & \cmark & \xmark & 970K trajectories & 2024-01 & 64xA100 & 14d & \cmark & \cmark \\
TinyVLA \citep{wen2025tinyvla} & SEP-2024 & 0.07/1.4 & Pythia/LLaVA & \cmark & \xmark & No large pre-training & 2024-07 & - & - & \cmark & \cmark \\

$\pi_0$ \citep{black2024pi_0} & OCT-2024 & 3.3 & PaliGemma & \cmark & \xmark & - & - & - & - & \cmark & \cmark \\
CogACT \citep{li2024cogact} & NOV-2024 & 7.3 & Prismatic-7B & \cmark & \xmark & 0.4M trajectories & - & 16×A100 & 5d & \cmark & - \\

$\pi_0$-FAST \citep{pertsch2025fast} & JAN-2025 & 3.3 & PaliGemma & \cmark & \xmark & 1M trajectories & - & - & 5× faster  $\pi_0$  & \cmark & \cmark \\

SpatialVLA \citep{qu2025spatialvla} & JAN-2025 & 4.03 & PaliGemma2-3B & \cmark & \xmark & 1.1M episodes & - & 64×A100 & 10d & \cmark & - \\

HiRobot \citep{shi2025hi} & FEB-2025 & ~3.3 & $\pi_0$ (PaliGemma) & \cmark & \xmark & - & - & - & - & \cmark & \cmark \\

GR00T-N1 \citep{bjorck2025gr00t} &MAR-2025 &2 &NVIDIA Eagle-2 VLM &\cmark &\xmark &592.9M trajectories &2025-02 & 1024 GPUs &5d &\cmark &\cmark\\

$\pi_{0.5}$ \citep{intelligence2025pi_} & APR-2025 & ~3.3 & $\pi_0$ (PaliGemma) & \cmark & \xmark & 1M traj+400h video & - & - & - & \cmark & \cmark \\

ChatVLA-2 \citep{zhou2025vision} & MAY-2025 & ~3-7 & Qwen-VL & \cmark & \xmark & VQA+Robotics & - & - & - & \cmark & \cmark \\

DriveMoE \citep{yang2025drivemoe} & MAY-2025 & ~3.3 & Drive-$\pi_0$ (PaliGemma) & \cmark & \xmark & Bench2Drive-4TB & - & - & - & \cmark & \cmark \\

DiffVLA \citep{jiang2025diffvla} & MAY-2025 & ~7B & Vicuna-v1.5-7B & \cmark & \xmark & - & - & - & - & \cmark & \cmark \\

WorldVLA \citep{cen2025worldvla} & JUN-2025 & ~7B & Chameleon & \cmark & \xmark & LIBERO-35GB & - & 200 GPUs & - & \cmark & \cmark \\

SmolVLA \citep{shukor2025smolvla} & JUN-2025 & 0.45 & SmolVLM-2 & \cmark & \xmark & LeRobot community & - & 1 GPU & - & \cmark & \cmark \\

AutoVLA \citep{zhou2025autovla} & JUN-2025 & 3 & Qwen2.5-VL-3B & \cmark & \cmark & - & - & - & - & \cmark & \cmark \\

DreamVLA \citep{zhang2025dreamvla} & JUL-2025 & 1.6 &  GPT-2 Medium & - & - & 1.75TB & - &  8xA800 & - & \cmark & \cmark \\

RynnVLA-001 \citep{RynnVLA2025} & AUG-2025 & 7 & Chameleon & - & - & 12M manipulation videos & - & - & - & \cmark & \cmark \\

\bottomrule
\end{tabular}%}
\end{table*}

\begin{table*}[!ht]
\centering
\caption{VLN historian}
\label{tab:model_adaptation VLN}
\tiny  % 最小字体
\setlength{\tabcolsep}{1.5pt}  % 减小列间距
\renewcommand\arraystretch{0.8}
%\resizebox{\textwidth}{!}{
\begin{tabular}{@{}lccccccccccc@{}} 
\toprule
& \multicolumn{8}{c}{Adaptation} & & \multicolumn{2}{c}{Evaluation} \\ % 合并第2到第9列
\cmidrule(lr){2-10}

\cmidrule(lr){11-12}

Model & \begin{tabular}[c]{@{}c@{}}Release \\ Time\end{tabular} & \begin{tabular}[c]{@{}c@{}}Size \\ (B)\end{tabular} & \begin{tabular}[c]{@{}c@{}}Base \\ Model\end{tabular} & \begin{tabular}[c]{@{}c@{}}IT \end{tabular} & \begin{tabular}[c]{@{}c@{}}RLHF \end{tabular} &  \begin{tabular}[c]{@{}c@{}}Pre-train \\ Data\end{tabular} & \begin{tabular}[c]{@{}c@{}}Latest Data \\ Timestamp\end{tabular} & \begin{tabular}[c]{@{}c@{}}Hardware \\ (GPUs / TPUs)\end{tabular} & \begin{tabular}[c]{@{}c@{}}Training \\ Time\end{tabular} & ICL   & CoT \\ 
\midrule
SayCan \citep{ahn2022can} & APR-2022 & 540 & PaLM/FLAN & \cmark & - & - & - & - & - & \cmark & \cmark \\
SSNet \citep{zhao2022zero} & JUN-2022 & - & existing LLM & - & - & - & - & - & - & \cmark & - \\
LM-Nav \citep{shah2023lm} & JUL-2022 & - & GPT-3/CLIP/ViNG & \cmark & - & - & - & - & - & \cmark & - \\
NavGPT \citep{zhou2024navgpt} & MAY-2023 & - & GPT-4/GPT-3.5 & \cmark & \cmark & - & - & - & - & \cmark & \cmark \\
YouTube-VLN \citep{lin2023learning} & JUL-2023 & - & VilBERT & \cmark & - & 1.2M videos & - & - & - & \cmark & - \\
PixNav  \citep{cai2024bridging} & SEP-2023 & - & GPT-4V/GroundingDINO & \cmark & - & - & - & - & - & \cmark & \cmark \\
VLN-CE \citep{xu2023vision} & OCT-2023 & - & LLM/VLM/DD-PPO & \cmark & - & - & - & - & - & \cmark & - \\
NaVid \citep{zhang2024navid} & FEB-2024 & - & Vicuna-7B & \cmark & \xmark & 510k video+763k traj& 2023-02 & - & - & \cmark & \cmark \\
VLN-Video \citep{li2024vln} & FEB-2024 & - & - & \cmark & \xmark & Touchdown+driving videos & 2024-01 & - & - & \cmark & \cmark \\
NaVILA \citep{cheng2024navila} & DEC-2024 & 8 & VILA/NVILA & \cmark & \xmark & - & 2024-12 & RTX 4090 & - & \cmark & \cmark \\
LH-Nav \citep{song2025towards} & DEC-2024 & - & - & \cmark & \xmark & LHPR-VLN 3260 tasks & 2024-12 & - & - & \cmark & \cmark \\
MapNav \citep{zhang2025novel} & FEB-2025 & - & VLM+SigLIP & \cmark & \xmark & ASM & 2025-01 & - & - & \cmark & \cmark \\
FlexVLN \citep{zhang2025flexvln} & MAR-2025 & - & LLM+Instruction Follower & \cmark & \xmark & Multi-dataset & 2025-03 & - & - & \cmark & \cmark \\
UniGoal \citep{yin2025unigoal} & MAR-2025 & - & LLM+Graph & \xmark & \xmark & Multi-goal & 2025-03 & - & - & \cmark & \cmark \\
WMNav \citep{nie2025wmnav} & MAR-2025 & - & Gemini VLM & \xmark & \xmark & World Model & 2025-03 & - & - & \cmark & \cmark \\
Dynam3D \citep{wang2025dynam3d} & MAY-2025 & 3.8 & LLaVA-Phi-3-mini+CLIP & \cmark & \xmark & 3D Multi-level & 2025-04 & RTX 6000 Ada & 9d & \cmark & \cmark \\
CityNavAgent \citep{zhang2025citynavagent} & MAY-2025 & - & existing LLM & - & - & - & - & - & - & \cmark & \cmark \\

\bottomrule
\end{tabular}%
%}
\end{table*}

\begin{table*}[!ht]
\centering
\caption{VLM historian}
\label{tab:model_adaptation VLM}
\tiny  % 最小字体
\setlength{\tabcolsep}{1.0pt}  % 减小列间距
\renewcommand\arraystretch{0.8}
%\resizebox{\textwidth}{!}{%
\begin{tabular}{@{}lccccccccccc@{}} 
\toprule
& \multicolumn{8}{c}{Adaptation} & & \multicolumn{2}{c}{Evaluation} \\ % 合并第2到第9列
\cmidrule(lr){2-10}

\cmidrule(lr){11-12}

Model & \begin{tabular}[c]{@{}c@{}}Release \\ Time\end{tabular} & \begin{tabular}[c]{@{}c@{}}Size \\ (B)\end{tabular} & \begin{tabular}[c]{@{}c@{}}Base \\ Model\end{tabular} & \begin{tabular}[c]{@{}c@{}}IT \end{tabular} & \begin{tabular}[c]{@{}c@{}}RLHF \end{tabular} &  \begin{tabular}[c]{@{}c@{}}Pre-train \\ Data\end{tabular} & \begin{tabular}[c]{@{}c@{}}Latest Data \\ Timestamp\end{tabular} & \begin{tabular}[c]{@{}c@{}}Hardware \\ (GPUs / TPUs)\end{tabular} & \begin{tabular}[c]{@{}c@{}}Training \\ Time\end{tabular} & ICL   & CoT \\ \\
\midrule
MOSAIC \cite{mandi2022towards} & OCT-2021 & - & existing VLM & - & - & - & - & - & - & \xmark & \xmark \\
language-planner \cite{huang2022language} & JAN-2022 & - & GPT-3/Codex & - & - & - & - & - & - & \cmark & \cmark \\
WHIRL \cite{bahl2022human} & JUL-2022 & - & existing VLM & - & - & - & - & - & - & \xmark & \xmark \\
MimicPlay \cite{wang2023mimicplay} & FEB-2023 & - & existing VLM & - & - & - & - & - & - & \xmark & \xmark \\
GD \cite{huang2023grounded} & MAR-2023 & - & existing VLM & - & - & - & - & - & - & \cmark & \cmark \\
ALOHA \cite{zhao2023learning} & APR-2023 & - & ACT/Transformer & - & - & - & - & - & - & \xmark & \xmark \\
VRB \cite{bahl2023affordances} & APR-2023 & - & CV algorithm & - & - & - & - & - & - & \xmark & \xmark \\
Statler \cite{yoneda2024statler} & JUN-2023 & - & existing VLM & - & - & - & - & - & - & \cmark & \cmark \\
RoboCLIP \cite{sontakke2023roboclip} & OCT-2023 & - & S3D/VLM & - & - & - & - & - & - & \xmark & \xmark \\
UniSim \cite{yang2023learning} & OCT-2023 & - & Diffusion/T5 & - & - & - & - & - & - & \xmark & \xmark \\
M-ALOHA \cite{fu2024mobile}  & JAN-2024 & - & ACT/Transformer & - & - & - & - & - & - & \xmark & \xmark \\
AlanaVLM \cite{suglia2024alanavlm} & JUN-2024 & 7 & Chat-UniVi & \cmark & - & EVUD & - & - & - & \cmark & \cmark \\
Grounded-VideoLLM \cite{wang2024grounded} & OCT-2024 & 4/8 & Phi3.5-Vision/LLaVA-Next & \cmark & - & Video Datasets & - & - & - & \cmark & \cmark \\
SeeDo \cite{wang2024grounded}& OCT-2024 & - & GPT-4o/VLM & - & - & Video Demos & - & - & - & \cmark & \cmark \\
RoboSpatial \cite{song2025robospatial} & NOV-2024 & - & existing VLM & \cmark & - & 1M images/5K scans & - & - & - & \xmark & \xmark \\
Video-3D LLM \cite{zheng2025video} & DEC-2024 & - & existing VLM & \cmark & - & 3D Datasets & - & - & - & \cmark & \cmark \\
OmniManip \cite{pan2025omnimanip} & JAN-2025 & - & - & \xmark & \xmark & - & 2025-01 & - & - & \cmark & \cmark \\
IKER \cite{patel2025real}& FEB-2025 & - & existing VLM & \xmark & \xmark & - & 2025-02 & - & - & \cmark & \cmark \\
RoboDexVLM \cite{liu2025robodexvlm} & MAR-2025 & - &  existing VLM & \xmark & \xmark & - & 2025-03 & - & - & \cmark & \cmark \\
VLM4D(benchmark) \cite{zhou2025vlm4d} & AUG-2024 & - & - & \xmark & \xmark & - & 2024-08 & - & - & \xmark & \xmark \\
VLM-3D \cite{chang2025vlm} & AUG-2025 & - & Qwen2-VL & \cmark & \xmark & - & 2025-08 & - & - & \cmark & \cmark \\

\bottomrule
\end{tabular}
%}
\end{table*}

\begin{table*}[!ht]
\centering
\caption{Agent historian}
\label{tab:model_adaptation Agent}

\tiny  % 最小字体
\setlength{\tabcolsep}{1.3pt}  % 减小列间距
\renewcommand\arraystretch{0.8}

\begin{tabular}{@{}lccccccccccc@{}} 

\toprule
& \multicolumn{8}{c}{Adaptation} & & \multicolumn{2}{c}{Evaluation} \\ % 合并第2到第9列
\cmidrule(lr){2-10}
\cmidrule(lr){11-12}

Model & \begin{tabular}[c]{@{}c@{}}Release \\ Time\end{tabular} & \begin{tabular}[c]{@{}c@{}}Size \\ (B)\end{tabular} & \begin{tabular}[c]{@{}c@{}}Base \\ Model\end{tabular} & \begin{tabular}[c]{@{}c@{}}IT \end{tabular} & \begin{tabular}[c]{@{}c@{}}RLHF \end{tabular} &  \begin{tabular}[c]{@{}c@{}}Pre-train \\ Data\end{tabular} & \begin{tabular}[c]{@{}c@{}}Latest Data \\ Timestamp\end{tabular} & \begin{tabular}[c]{@{}c@{}}Hardware \\ (GPUs / TPUs)\end{tabular} & \begin{tabular}[c]{@{}c@{}}Training \\ Time\end{tabular} & ICL   & CoT \\ \\
\midrule
GATO \citep{reed2022generalist} & MAY-2022 & 1.18 & - & \cmark & \xmark & - & 2022-05 & - & - & \cmark & \cmark \\
Inner Monologue \citep{huang2022inner} & JUL-2022 & - & existing LLM & \xmark & \xmark & - & 2022-07 & - & - & \cmark & \cmark \\
ProgPrompt \citep{singh2022progprompt} & SEP-2022 & - & existing LLM & \xmark & \xmark & - & 2022-09 & - & - & \cmark & \cmark \\
LLM-Planner \citep{song2023llm} & DEC-2022 & - & GPT-3 & \xmark & \xmark & - & 2022-12 & - & - & \cmark & \cmark \\
Voyager \citep{wang2023voyager} & MAY-2023 & - & GPT-4 & \xmark & \xmark & - & 2023-05 & - & - & \cmark & \cmark \\
RoboCat \citep{bousmalis2023robocat} & JUN-2023 & - & Gato & \cmark & \xmark & - & 2023-06 & - & - & \cmark & \cmark \\
RoboAgent \citep{bharadhwaj2024roboagent} & SEP-2023 & - & - & \xmark & \xmark & 7500 trajectories & 2023-09 & - & - & \cmark & \cmark \\
LEO \citep{huang2023embodied}& NOV-2023 & - & Vicuna-7B & \cmark & \xmark & - & 2023-11 & - & - & \cmark & \cmark \\
RoboFlamingo \citep{li2023vision} & NOV-2023 & - & OpenFlamingo & \cmark & \xmark & - & 2023-11 & - & - & \cmark & \cmark \\

ELLMER \citep{mon2025embodied} & JUN-2024 & - & GPT-4 & \xmark & \xmark & - & 2024-06 & - & - & \cmark & \cmark \\
ECoT \citep{zawalski2024robotic} & JUL-2024 & - & OpenVLA & \cmark & \xmark & - & 2024-07 & - & - & \cmark & \cmark \\
SPINE \citep{ravichandran2025spine} & OCT-2024 & - & existing LLM & - & - & - & - & - & - & - & -\\
RoboMatrix \citep{mao2024robomatrix} & DEC-2024 & - & LLaVA & \cmark & - & 1.5K videos & - & 8xA100 & - & \cmark & \cmark \\
GeneWorker \citep{wang2024geneworker} & OCT-2024 & - & existing LLM & - & - & - & - & - & - & - & - \\
Mem2Ego \citep{zhang2025mem2ego} & FEB-2025 & 11 & Llama3.2-11B & \cmark & - & - & - & 24GB+ GPU memory & - & \cmark & \cmark \\
RoboFlamingo-Plus \citep{wang2025roboflamingo} & MAR-2025 & 3-9 & OpenFlamingo & \cmark & - & - & - & 64xA100 & - & \cmark & \cmark \\
LLM+MAP \citep{chu2025llm+} & MAR-2025 & - & GPT-4o & - & - & - & - & - & - & \cmark & \cmark \\
Agentic Robot \citep{yang2025agentic} & MAY-2025 & 7 & OpenVLA-7B & \cmark & - & LIBERO-35GB & - & - & - & \cmark & \cmark \\
FLARE \citep{zheng2025flare}& MAY-2025 & - & Diffusion Transformer & - & - & - & - & - & - & \cmark & \cmark \\

\bottomrule

\end{tabular}
\end{table*}

\subsection{Popular LLMs Companies}

This section provides an overview of prominent companies at the forefront of LLM development. We present concise profiles of these industry leaders and introduce their latest model series.

\textbf{OpenAI}, a pioneering AI research and deployment organization, is dedicated to harnessing the power of general AI to benefit humanity as a whole while catalyzing a surge in LLM research and development. The GPT series is one of the most popular LLM today, encompassing notable iterations such as GPT-3, GPT-3.5-Turbo, GPT-4 family, OpenAI o1, and etc \citep{zhao2023survey}.  Recently, GPT-5 \citep{GPT5} represents OpenAI's most advanced general-purpose AI model to date, incorporating a Mixture-of-Experts (MoE) architecture. GPT-5 combines the linguistic capabilities of OpenAI's GPT-series with the analytical reasoning strengths of their O-series models and features native multimodal processing capabilities. Concurrently, OpenAI has introduced the GPT-OSS series \citep{GPTOSS}, an open-source collection of LLMs released under permissive commercial licensing. These models employ MoE architecture specifically optimized for complex reasoning tasks and enhanced tool manipulation capabilities, while maintaining full transparency and modifiability for community development.

\textbf{Meta} focuses on bringing the metaverse to life, aiming to connect people, foster community growth, and enable businesses to thrive. Additionally, the company is actively researching and developing LLMs. LLaMA series is one of the most popular open-sources, boasting a vibrant open-source ecosystem and community. Notably, it prioritizes compatibility with open-source frameworks and emphasizes reproducibility, ensuring its wide adoption and development. LLama's training data is sourced from publicly available datasets, without relying on proprietary or customized datasets. The Llama series of LLMs includes Llama-2, Llama-3, and Llama-4 series \citep{zhao2023survey, meta2025llama}. Meta's most recent advancement, DINOv3 \citep{simeoni2025dinov3}, represents a state-of-the-art (SOTA) visual model that synergistically combines elements of the Vision Transformer (ViT) and ConvNeXt architectures. Remarkably, DINOv3 achieves performance on par with, or even surpasses specialized models across multiple computer vision benchmarks.

\textbf{Google}, a multinational technology company, has also made significant contributions to the development of LLMs; it is indeed a pioneer in advancing large-scale language modeling. Over the years, Google has introduced a variety of LLM architectures targeting different domains, such as BERT, T5, and PaLM \citep{zhao2023survey}. In particular, Google's DeepMind division has been instrumental in bridging LLMs with broader artificial general intelligence (AGI) research. DeepMind Lab has long served as a key experimental platform for embodied intelligence, multi-agent coordination, and reinforcement learning, providing foundational insights that have guided the design of reasoning and planning components in later-generation LLMs. Additionally, \textbf{Gemini} (originally named Bard) \citep{comanici2025gemini} stands as one of Google’s flagship multimodal models. It integrates advances from both Google Research and DeepMind, combining large-scale language understanding with vision and reasoning capabilities. Gemini is released in three distinct scales—\textit{Ultra}, \textit{Pro}, and \textit{Nano}—each optimized for different application contexts: \textit{Ultra} targets complex reasoning and multimodal analysis, including visual-text understanding; \textit{Pro} supports general-purpose tasks; and \textit{Nano} is designed for efficient on-device deployment. More recently, \textbf{Genie 3} \citep{genie3} has emerged as Google’s most advanced generative system and the first demonstrated universal world model capable of real-time interactive simulation. Building on the foundation of Gemini and DeepMind’s long-standing research on grounded intelligence, Genie 3 can dynamically synthesize explorable 3D environments from textual prompts while maintaining long-term visual and physical consistency. This integration highlights Google’s strategic move toward unifying large language modeling, embodied reasoning, and simulation-based intelligence under a cohesive framework.

\textbf{Anthropic}, a cutting-edge AI startup, was co-founded by a team of renowned experts, including the former Vice President of OpenAI and the lead author of the seminal GPT-3 paper, among other notable individuals. Claude is a suite of high-performance, intelligent AI models developed by Anthropic, designed to push the boundaries of language understanding and generation capabilities. Claude is designed to adhere to key protocols, minimize errors, and resist jailbreak attacks, empowering enterprise customers to develop and deploy highly secure, AI-driven applications at scale. The Claude series of LLMs comprises Claude-2 and the Claude-3 family \citep{zhao2023survey}. The latest addition to the family is Claude 4.1 Opus and Claude Sonnet 4. Claude 4.1 Opus \citep{ClaudeOpus} represents Anthropic's current flagship model, specifically optimized for three demanding computational domains: complex programming logic, multi-agent system coordination, and deep analytical reasoning. The model features an industry-leading 200K token context window, enabling sophisticated capabilities such as cross-file codebase refactoring and large-scale software project navigation. Claude Sonnet 4 \citep{ClaudeSonnet4} represents an optimized balance between computational efficiency and model performance, specifically designed for high-throughput applications including document processing and automated code debugging. 

\textbf{DeepSeek} is a pioneering AI startup. The company specializes in developing cutting-edge LLMs and has swiftly established itself as a key player in the global AI landscape. DeepSeek has unveiled a suite of LLM series, spanning a wide range of applications, including general dialogue, code generation, mathematical reasoning, and multimodal understanding, among others.  
In terms of reasoning, the DeepSeek-R series \citep{deepseekai2025deepseekr1incentivizingreasoningcapability} leverages reinforcement learning to enhance its inferential capabilities, with a focus on logical reasoning. The DeepSeek-Math \citep{deepseek-math} series excels in mathematical reasoning tasks, demonstrating exceptional performance across a range of prestigious mathematical benchmark tests. In terms of general dialogue, the DeepSeek-V \citep{deepseekai2024deepseekv3} series excels in generating human-like content and engaging in conversations across diverse topics, with a strong emphasis on generality and advanced natural language processing capabilities. In terms of MLLMs, DeepSeek VL \citep{wu2024deepseekvl2} is a multimodal model that seamlessly integrates visual and linguistic information, enabling comprehensive understanding and generation of both images and text. In terms of code generation, DeepSeek Coder \citep{zhu2024deepseek} is a cutting-edge model designed for generating and understanding code, with support for a wide range of programming languages.

\textbf{Alibaba} has established itself as a preeminent e-commerce and technology conglomerate. Originally founded as an e-commerce platform, the corporation has since evolved into a multinational technology enterprise with diversified business segments encompassing cloud computing services, digital financial ecosystems, intelligent logistics networks, and advanced AI research. Qwen represents a series of LLMs developed by Alibaba Group DAMO Academy as part of its proprietary generative AI technologies. As a comprehensive foundation model series, Qwen demonstrates state-of-the-art performance across multiple natural language processing tasks. Qwen2.5-Max \citep{qwen25} possesses multimodal capabilities, including advanced visual understanding for specialized domains such as medical image diagnostics. The model has been successfully deployed across Alibaba's intelligent service ecosystem, powering both customer service automation and AI-driven content generation platforms. Qwen3 \citep{qwen3} represents the most advanced base model in the Qwen series, incorporating a novel MoE architecture. As the first LLM globally to implement dual-mode 'fast thinking' and 'slow thinking'  with seamless transition capabilities. Qwen3-Coder \citep{qwen3technicalreport} leverages the MoE architecture inherited from Qwen3, specializing in programming-related tasks. This domain-specific variant demonstrates exceptional capabilities in four key areas: automated code generation,  program repair and optimization, intelligent debugging assistance, and agent-based programming workflows.

\begin{figure}
    \centering
    \includegraphics[scale=0.8]{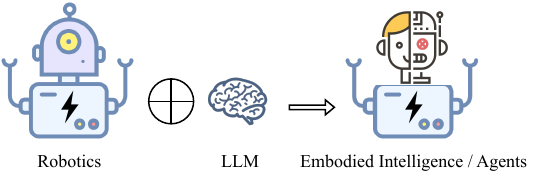}
    \caption{Robotics based on LLM.}
    \label{fig: robotics combined with LLMs}
\end{figure}

\section{Robotics Based on LLMs}  
\label{sec: LLM for Robotics}

In this section, we introduce some representative robot transformer foundation models nowadays. Additionally, this section introduces robotics based on LLMs, including current representative agents and embodied robotics. LLMs are used as brains in robotics, like in Fig. \ref{fig: robotics combined with LLMs}.

\subsection{Agent AI}
\begin{figure}
    \centering
    \includegraphics[scale=0.42]{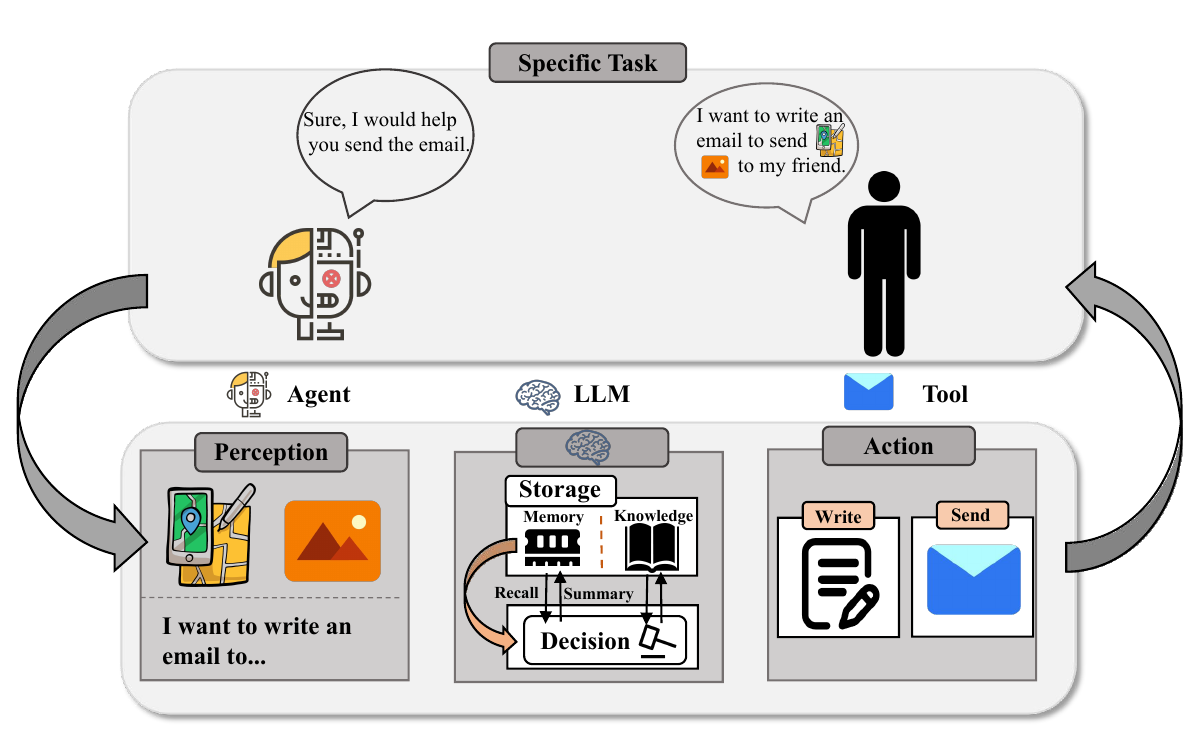}  
     \caption{The conceptual framework of an LLM-based agent comprises three components: brain, perception, and action. The brain module serves as the controller, responsible for core tasks such as memory, reasoning, and decision-making. The perception module captures and processes multimodal information from the external environment, while the action module executes tasks using tools and interacts with the surroundings.}
    \label{fig: agent}
\end{figure}

The agent is a sophisticated system that possesses the ability to perceive its environment, make informed decisions, and execute actions to achieve specific goals. The Agent AI technology is currently at a pivotal juncture, undergoing a significant transformation from `intelligent assistants' to `autonomous agents'. This shift is being propelled by the convergence of embodied AI, LLMs, and advanced control, perception, and action modules. As a result, AI agents are rapidly evolving to possess more general, task-oriented, and realistic intelligent capabilities. The basic structure of the agent is illustrated in Fig.~\ref{fig: agent}.

\subsubsection{Multi-Agent}

\begin{figure}
    \centering
    \includegraphics[scale=0.56]{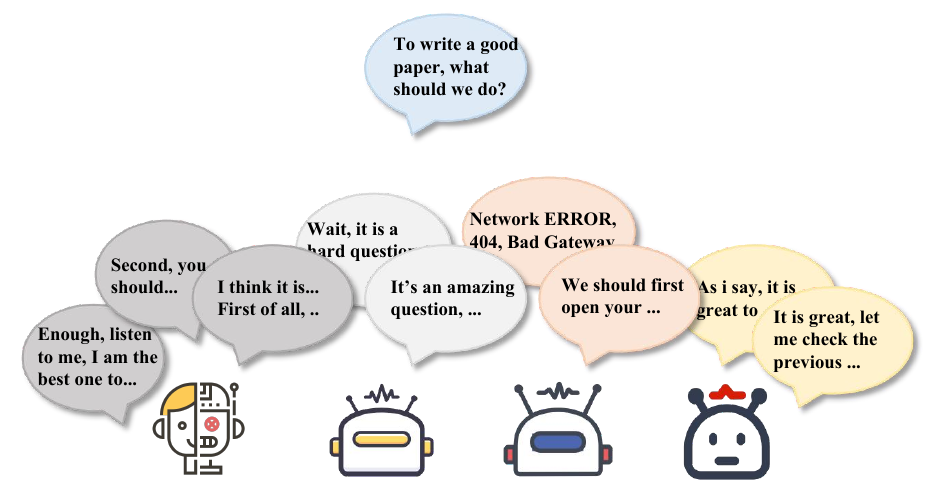}  
     \caption{Scenarios of interaction among multiple LLM-based agents.}
    \label{fig: multiagent}
\end{figure}

Grounding the reasoning capabilities of LLMs in embodied tasks is a challenging endeavor, owing to the inherent complexity and nuances of the physical world. LLMs for planning in multi-agent collaboration necessitate effective communication among agents and credit assignment as feedback mechanisms to iteratively refine proposed plans and attain seamless coordination. The example of the multi-agent system is presented in Fig. \ref{fig: multiagent}. Current LLM-based multi-agent systems (LLM-based MAS) have demonstrated significant success in various applications. The tasks of Multi-Agent Systems (MAS) can be broadly categorized into two types based on the physical structure of the robots: (1) collaboration among multiple isomorphic robots, and (2) collaboration among multiple heterogeneous robots.

For multiple isomorphic robots, RoCo \citep{mandi2024roco} leverages the capabilities of LLMs to facilitate both collaboration and task decomposition in multi-arm motion. Furthermore, it introduces RoCoBench, a comprehensive 6-task benchmark for multi-robot manipulation, which will be made openly available to the research community. LLM-Debate \citep{du2023improving} presents a complementary approach to enhancing language responses, wherein multiple instances of language models engage in a multi-round debate, proposing and refining their individual responses and reasoning processes to converge on a consensus final answer. 

For multiple heterogeneous robots, COHERENT \citep{liu2025coherent} is an LLM-based task planning framework designed to facilitate collaboration among heterogeneous multi-robot systems, comprising quadrotors, robotic dogs, and robotic arms. The proposed framework features a task assigner and a robot executor that decomposes human instructions into subgoals, which can be assigned to specific robots. EMOS \citep{chen2024textbf} is an LLM-based multi-agent system specifically designed to control and coordinate multi-robot systems in complex household environments. By facilitating effective collaboration among heterogeneous robots with diverse embodiments and capabilities, EMOS enables seamless interaction and task execution. Additionally, EMOS introduces a new benchmark, Habitat-MAS.

Others do not specifically refer to isomorphism or heterogeneity. ReAd \citep{zhang2024towards} enables efficient self-refinement of plans by performing critic regression to learn a sequential advantage function from data generated by the LLM planner. This learned function is then used to treat the LLM planner as an optimizer, which generates actions that maximize the advantage function. LLaMAC \citep{zhang2023controlling} is a modular framework designed to mitigate hallucinations in LLMs and improve coordination in MAS as the number of agents scales up. By establishing an external feedback mechanism between LLM-based actors and the TripletCritic, LLaMAC enhances the collaborative performance of large-scale multi-agent systems. This study \citep{chen2024scalable} compares the task success rate and token efficiency of four distinct multi-agent communication frameworks - centralized, decentralized, and two hybrid approaches - across four coordination-dependent multi-agent 2D task scenarios, with a focus on evaluating their performance as the number of agents increases.

\subsubsection{Agent Framework}
Long-horizon robotic manipulation presents a formidable challenge for autonomous systems, necessitating prolonged reasoning, precise execution, and robust error recovery across intricate sequential tasks. Existing approaches are hindered by error accumulation and a lack of effective verification mechanisms during execution. Currently, several frameworks have been proposed to address this challenge. REMAC \citep{yuan2025remac} employs a 'self-reflection and self-evolution' mechanism, which enables continuous pre-condition and post-condition checks, as well as real-time re-planning in response to failure situations. This adaptive approach supports collaboration among multiple robots; Recoverychaining \citep{vats2024recoverychaining} employs a hierarchical strategy to initiate local corrections of the Recovery Policy upon detecting execution failures; While Agentic Robot \citep{yang2025agentic} is a notable and representative example among them.

Agentic Robot \citep{yang2025agentic}, a brain-inspired framework, overcomes these limitations by introducing the Standardized Action Procedure (SAP) - a novel coordination protocol that governs component interactions throughout manipulation tasks. Inspired by Standardized Operating Procedures (SOPs) used in human organizations, the proposed SAP establishes a structured workflow that spans planning, execution, and verification phases. An Agentic Robot consists of three specialized components. These components include: (1) a large-scale reasoning model that decomposes high-level instructions into semantically coherent subgoals; (2) a VLA executor that generates continuous control commands from real-time visual inputs; and (3) a temporal verifier that facilitates autonomous progression and error recovery through introspective assessment. The SAP-driven closed-loop architecture enables dynamic self-verification. Additionally, Agentic Robot achieves state-of-the-art (SOTA)  average success rate on the LIBERO benchmark.

\subsection{Embodied Robotics}

Embodied Intelligence in Humanoid Robots seeks to equip robots with a human-like body structure and endow them with intelligence, enabling them to perceive, understand, reason, and interact with their physical environment in a manner akin to humans. Numerous companies have recently unveiled their humanoid robots \citep{g1, atlas}. In this section, we will showcase some of the most notable examples. Concurrently, we have compiled a list of commonly used humanoid datasets in Table \ref{table:dataset_stats} for reference. The Table is cited from \citep{bjorck2025gr00t}. Furthermore, we have also summarized the benchmarks of Libero \citep{liu2023libero} in Table \ref{table:libero_series}, providing a comprehensive overview. Note that the dataset size listed in Table \ref{table:libero_series} is the size of the compressed file.

\textbf{UniTree,} is a leading robotics company, renowned for developing high-performance, affordable quadruped and humanoid robots. With a mission to bridge the gap between research and reality, Unitree is dedicated to empowering robots to transition from laboratory settings to real-world applications. Unitree's product portfolio comprises two primary series: quadruped robots and humanoid robots. The quadruped robot series can be further categorized into two main lines: the Go series and the B series. The Go series serves as Unitree's entry-level quadruped robot, primarily designed for scientific research and educational purposes \citep{go1, go2}. In contrast, the B series is a high-performance, industrial-grade quadruped robot, optimized for applications in industrial inspection, logistics transportation, agricultural automation \citep{b1, b2}, etc. Unitree's humanoid robots can be divided into two distinct series: the H series and the G series. The H series represents Unitree's first-generation humanoid robot, boasting advanced capabilities such as bipedal self-balancing, stable running, and motion imitation \citep{hseries}. Equipped with a comprehensive array of IMU and encoder sensors throughout its body, the H series is well-suited for research and industrial applications. The G series, on the other hand, is Unitree's flagship line of general-purpose humanoid robots, with the ``G'' designation symbolizing ``General Purpose'' or ``General Intelligence". This series is designed to achieve a harmonious integration of humanoid perception, cognition, and motor abilities, enabling it to meet the diverse demands of various tasks in real-world environments \citep{g1, g1comp}. By doing so, the G series aims to push the boundaries of humanoid robotics and unlock new possibilities for applications in multiple industries.

\textbf{Boston Dynamics} is a world-renowned robotics company celebrated for its innovative, biomimetic robot designs that emulate the agility and adaptability of humans and animals. Focused on pushing the boundaries of robotics, Boston Dynamics is dedicated to creating robots that can perceive, move, and adapt in a manner indistinguishable from living creatures. Boston Dynamics' product portfolio can be broadly categorized into three main types: the quadruped robot Spot, the logistics robot Stretch, and the humanoid robot Atlas. Spot, an agile "robotic dog," is designed for complex environmental inspections. It can be equipped with various modules, including LiDAR, cameras, and robotic arms. It supports autonomous navigation, remote control, and multi-robot formation capabilities. Its primary applications include inspecting hazardous areas in petroleum, chemical, and power plants, as well as providing reconnaissance support for public security and rescue missions \citep{spot}. Stretch, a logistics automation robot, is designed to serve the warehousing and logistics industries. Unlike traditional solutions, it does not require fixed infrastructure and can be deployed in standard warehouses. Its primary function is to automate manual, repetitive, and labor-intensive tasks \citep{stretch}. Atlas, a research platform for humanoid dynamic robots, is capable of complex movements, including running, jumping, and backflips on two feet, showcasing advanced dynamic coordination \citep{atlas}. Currently, Atlas is a non-commercial product used exclusively for cutting-edge research and demonstrations, pushing the boundaries of robotics and AI.

\textbf{Figure AI}. Figure AI's flagship product is the Figure series of humanoid robots, a groundbreaking ``universal humanoid robot'' developed in collaboration with OpenAI \citep{figure}. This innovative robot integrates OpenAI's multimodal models, enabling advanced capabilities such as visual understanding, language interaction, and reasoning. One of the key features of the Figure series is its ability to learn new tasks autonomously through imitation demonstrations, such as preparing coffee or cleaning dishes. Additionally, the robot's scene voice interaction capability allows it to describe its surroundings, provide real-time feedback on task status, and engage in conversational dialogue. The Figure series is primarily designed to automate repetitive and mundane tasks, freeing humans from tedious labor. Its applications include box handling, sorting tasks, cleaning, organizing items, and providing daily assistance, among others.

\textbf{Tesla,} a global leader in electric vehicles and autonomous technologies, is expanding its innovation into humanoid robotics through the development of the \textit{Tesla Optimus} project \citep{Optimus}. Leveraging Tesla’s deep expertise in AI, computer vision, and real-world data processing, the company aims to create a general-purpose humanoid robot capable of performing a wide range of everyday tasks, bridging the gap between automation and human-level adaptability.

\textbf{Figma,} originally known for its collaborative design and prototyping platform, is expanding its innovation frontier into the field of intelligent design robotics \citep{figma}. Building on its strong foundation in human–computer interaction and real-time collaboration, Figma is developing a new generation of \textit{design-assist robots} that integrate multimodal intelligence with creative reasoning. These robots are envisioned to act as embodied creative partners, capable of transforming design concepts into tangible prototypes through intuitive dialogue and adaptive physical manipulation.

\textbf{Fourier Intelligence} is a pioneering global leader in rehabilitation and humanoid robotics. Fourier Intelligence's embodied robotic product portfolio encompasses two primary categories: the open-source N1 robot and the GR series. 
The Fourier N1 represents Fourier Intelligence's inaugural open-source humanoid robotics platform, featuring comprehensive disclosure of its ontological framework. This release includes a complete technical resource package encompassing bill of materials, mechanical design schematics, detailed assembly protocols, and foundational operating software \citep{fouriern12025}. The GR series encompasses three distinct robots: GR-1 and GR-2 serve as general-purpose robotic assistants \citep{fouriergr12023, fouriergr22024}, while GR-3 is a humanoid companion robot specifically designed for affective human-robot interaction. GR-3's core design philosophy emphasizes psychological affinity achieved through multimodal sensory integration and biomimetic expressiveness.

\textbf{AgiBot,} is a pioneering robot brand. With a strong commitment to advancing the fusion of AI and robotics, AgiBot is dedicated to the research, development, and production of cutting-edge, universal humanoid robots that set new standards for intelligent robotics worldwide \citep{agibotoffical}. They contributed the world's largest open-source humanoid robot dataset, agibot-world \citep{agibotworld}. Agibot is dedicated to developing humanoid robot platforms, with a primary focus on creating advanced humanoid robots. The company's flagship product, AgiBot A2 series, is specifically designed for industrial applications, offering a range of models to cater to diverse needs. The A2 is a bipedal humanoid robot, standing approximately 1.75m tall, equipped with autonomous navigation, language interaction, and environmental adaptation capabilities, making it suitable for general interaction and industrial tasks \citep{agibota2}.

The X series is a lightweight, interactive humanoid robot designed to provide emotional interaction and household assistance, making it perfect for family and community settings \citep{agibotx}. The G1, for instance, is a wheeled dual-arm robot tailored for warehousing and logistics, featuring dual arm and wheeled mobile modules that enable efficient cargo transportation and handling \citep{agibotg1}. The C5 is a versatile robot that supports integrated operations such as sweeping, washing, and dust pushing, equipped with advanced laser vision positioning, automatic water and sewage stations \citep{agibotc5}.

\section{Related Technologies} \label{sec:technologies}

\begin{figure*}
    \centering
    \subfigure[The robot uses its sensor to scan the environment, then calculates the quantity of the sheep in the grass.]{
        \label{fig:Perception}
        \includegraphics[scale = 0.48]{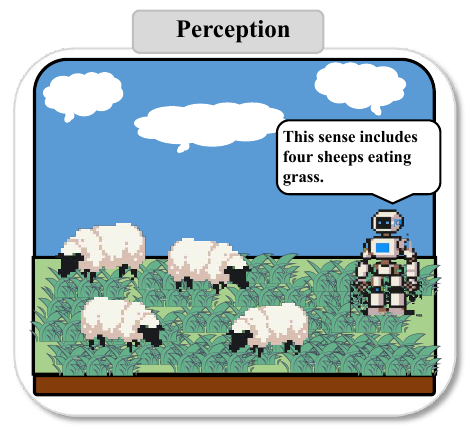}
    }\hspace{2mm}
    \subfigure[The robot meets an emergency, then quickly makes several decisions about how to handle it.]{
        \label{fig: Decision making}
        \includegraphics[scale = 0.5]{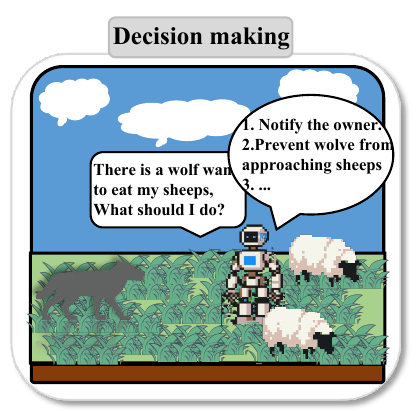}
    }\hspace{2mm}
    \subfigure[The robot picks up a weapon and want to against the wolf.]{
        \label{fig: Action}
        \includegraphics[scale = 0.5]{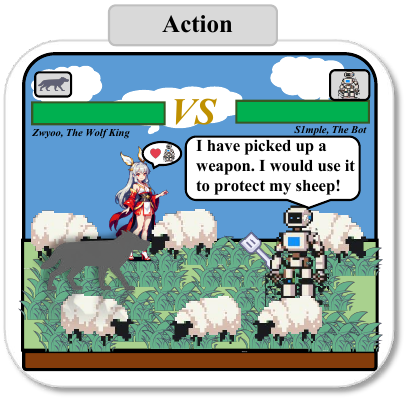}
    }\hspace{2mm}
    \subfigure[The robot makes a conversation with a bunny girl about how to do with the situation.]{
        \label{fig: Interaction}
        \includegraphics[scale = 0.5]{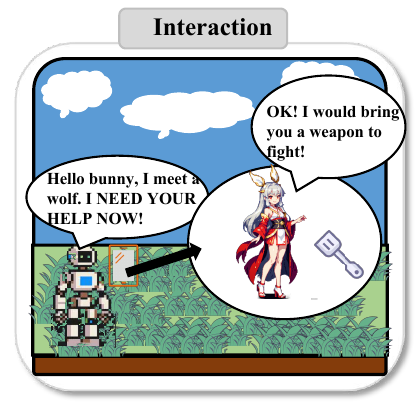}
    }
    \caption{The embodied robotics is mainly composed of four components, which are Perception, Decision-making, Action, and Interaction. Here, we show the ability of each components based on a little story.}
    \label{fig:embodied robotics}
\end{figure*}

In this section, we introduce the related technology used in robotics. Here, we divided the model of robotics into four parts, which are perception, decision-making, control, and interaction. We certainly provide a more detailed introduction to the decision-making component, as it serves as the core of robotics based on LLMs. Decision-making serves as a connecting link between perception and control. Additionally, we give a little story to illustrate the distinct components of embodied robots in Fig. \ref{fig:embodied robotics}. 

\begin{table*}[htbp]
\centering
\footnotesize
\caption{Pre-training dataset statistics}
\label{tab:Pre-training Dataset Statistics}
\begin{tabular}{llllll}
\toprule
Dataset & Length (Frames) & Duration (hr) & FPS & Camera View & Category \\ 
\midrule
GR-1 Teleop Pre-Training & 6.4M & 88.4 & 20 & Egocentric & Real robot \\
DROID (OXE) & 23.1M & 428.3 & 15 & Left, Right, Wrist & Real robot \\
RT-1 (OXE) & 3.7M & 338.4 & 3 & Egocentric & Real robot \\
Language Table (OXE) & 7.0M & 195.7 & 10 & Front-facing & Real robot \\
Bridge-v2 (OXE) & 2.0M & 111.1 & 5 & Shoulder, left, right, wrist & Real robot \\
MUTEX (OXE) & 362K & 5.0 & 20 & Wrist & Real robot \\
Plex (OXE) & 77K & 1.1 & 20 & Wrist & Real robot \\
RoboSet (OXE) & 1.4M & 78.9 & 5 & Left, Right, Wrist & Real robot \\
Agibot-Alpha & 213.8M & 1,979.4 & 30 & Egocentric, left, right & Real robot \\
RH20T-Robot & 4.5M & 62.5 & 20 & Wrist & Real robot \\
Ego4D & 154.4M & 2,144.7 & 20 & Egocentric & Human \\
Ego-Exo4D & 8.9M & 123.0 & 30 & Egocentric & Human \\
Assembly-101 & 1.4M & 19.3 & 20 & Egocentric & Human \\
HOI4D & 892K & 12.4 & 20 & Egocentric & Human \\
HoloAssist & 12.2M & 169.6 & 20 & Egocentric & Human \\
RH20T-Human & 1.2M & 16.3 & 20 & Egocentric & Human \\
EPIC-KITCHENS & 2.3M & 31.7 & 20 & Egocentric & Human \\
GR-1 Simulation Pre-Training & 125.5M & 1,742.6 & 20 & Egocentric & Simulation  \\
GR-1 Neural Videos & 23.8M & 827.3 & 8 & Egocentric & Neural-generated \\
\midrule

\end{tabular}
\label{table:dataset_stats}
\end{table*}

\begin{table*}[htbp]
\centering
\caption{Overview of LIBERO series datasets}
\begin{tabularx}{\textwidth}{X X X X}
\hline
\textbf{Dataset} & \textbf{Task Focus} & \textbf{Number of Tasks} & \textbf{Dataset Size} \\
\hline
LIBERO-SPATIAL & Spatial reasoning & 10 & 2.68 GB \\
LIBERO-OBJECT & Object manipulation & 10 & 3.97 GB \\
LIBERO-GOAL & Goal-conditioned tasks & 10 & 2.68 GB \\
LIBERO-100 & Long-horizon tasks & 100 & 31.64 GB \\
\hline
\end{tabularx}
\label{table:libero_series}
\end{table*}

\subsection{Perception}

Perception is a fundamental capability of robots, akin to their input. Currently, multi-modality is a popular approach for robot perception. The models discussed below employ different treatments of perception.

\subsubsection{Vision-Language Model}

In recent years, LLMs and visual models have achieved great success in their respective fields. However, each model can only process inputs in its corresponding domain (for example, the language model only accepts text as input, and the visual model only accepts images as input), which is relatively limited. Researchers have thus begun to focus on the processing of multi-modal input, combining LLMs and visual models. Therefore, the multi-modal model that can take both vision and natural language as input was created --- the visual-language model (VLM). VLM can process images and text at the same time. The development of VLMs has progressed through three major phases: Self-supervised pretraining, contrastive pretraining, and large multimodal models. In actual use, we also need to distinguish between recognizing 2D scenes (such as some Visual Transformers (ViTs) \citep{chen2022pali, dosovitskiy2020image, ryoo2021tokenlearner}) or 3D scenes (such as OSRT \citep{sajjadi2022object}) when processing vision. VLMs come in various types \citep{gan2022vision}. There are many VLM models emerging. Contrastive Language-Image Pre-training (CLIP) \citep{radford2021learning} is a neural network that has been trained on diverse pairs of images and text. It has the capability to understand natural language instructions and predict the most pertinent text excerpts associated with a given image, all without directly optimizing for this specific task. CLIP is similar to the zero-shot function of GPT-2 and 3. CLIP is also used in LM-Nav \citep{shah2023lm} as a VLM to predict the text from natural language instructions. The landmarks are extracted and incorporated into the topological map. VLMs are versatile and can be employed in various downstream tasks, including visual question answering (VQA) \citep{zellers2021merlot, zhou2020unified}, optical character recognition (OCR) \citep{li2023trocr}, and image captioning \citep{hu2022scaling}. Such as PaLM-E \citep{driess2023palm} treats text and images as latent vectors of multi-modal input. Frozen \citep{tsimpoukelli2021multimodal} is also processed similarly to PaLM-E. PhysVLM \citep{zhou2025physvlm} has pioneered the development of a novel Spatial Accessibility Map (S-P Map), which seamlessly integrates the actual operational range of robots into VLM inputs. This innovative approach yields significant performance enhancements in physical interaction tasks.

\subsubsection{Scene Understanding}

Scene Understanding encompasses a deep comprehension of the intricate relationships between objects, structures, and semantics within a given environment. This embodied intelligence, akin to the symbiotic functioning of the ``eyes" and the ``cerebral cortex", enables robots to develop a nuanced understanding of their surroundings, allowing them to make informed, context-dependent decisions that drive meaningful behavioral responses. The use of pretrained backbones with fine-tuning has been successful for 2D vision and natural language processing tasks, showing advantages over task-specific networks. Swin3D \citep{yang2023swin3d} is a pretrained 3D backbone for 3D indoor scene understanding. It includes a 3D Swin transformer as a backbone network, which enables efficient self-attention on sparse voxels with linear memory complexity, making the backbone scalable to large models and datasets. BIP3D \citep{lin2025bip3d} seamlessly integrates pre-trained 2D visual features with explicit 3D position encoding, achieving end-to-end prediction from multi-view images to 3D perception. Notably, this approach yields significant improvements on the EmbodiedScan benchmark. HumanoidPano \citep{zhang2025humanoidpano} integrating 360° panoramic vision with LiDAR,  leverages a spherical Transformer architecture to achieve precise image-depth alignment. This enables the generation of high-quality bird's-eye views, facilitating effective navigation tasks.

\subsubsection{Indirect Perception}

Indirect Perception is a paradigm that underscores the notion that perceptual outputs are not solely utilized for immediate action, but also inform higher-level cognitive processes such as planning, decision-making, and multi-step reasoning. This mechanism is crucial for enabling robots to adopt a ``think-before-acting'' approach, where they can develop a deeper understanding of their environment before executing actions. Compared with direct perception using visual modality, GRUtopia
\citep{wang2024grutopia} proposes an indirect perception method that uses LLMs to act as NPCs (Non-Player Characters) to provide robots with environmental knowledge and action feedback. GRUtopia contains GRResidents, an LLM-driven NPC system that is responsible for social interaction, task generation, and task assignment, thus simulating social scenarios for embodied AI applications. Visual affordances \citep{apicella2025visual} integrate multiple visual affordance tasks and propose a comprehensive perception framework that establishes a direct relationship between the visual and physical worlds, thereby facilitating a seamless bridge between perception and action.

\subsection{Decision-making}

Decision-making is a fundamental capability of robots, enabling them to make informed decisions and plan tasks based on their current state and environment. As the core of a robot, decision-making plays a crucial role in connecting the preceding and the following, analyzing input from the perception module to generate appropriate actions. Policy refers to a specific function or mechanism that enables decision-making, effectively mapping observations to actions through a well-defined relationship. In the context of robotics, policies play a crucial role in determining the actions an agent takes in response to its environment. Furthermore, with the increasing complexity of robotic systems, multi-robot collaboration has become a significant area of research, where individual or joint policy learning is used to enable coordinated decision-making among multiple agents. In the following content, we introduce the details of the policy and multi-robot collaboration.

\subsubsection{Decision Policy}

\textbf{What Brings Intelligence to Robotics}? LLMs have the potential to significantly aid intelligent agents, with numerous studies successfully utilizing LLMs as the brain to implement intelligent agents \citep{brohan2022rt, driess2023palm, shah2023lm} and achieve promising results \citep{ouyang2022training, radford2018improving}. Ideally, embodied intelligence should constitute an intelligent entity capable of perceiving its surrounding environment and producing appropriate outputs after interacting with humans or the environment. LLM plays a vital role in this process, serving as a central hub for analyzing multi-modal input and converting it into suitable action outputs. The development of intelligent agents has progressed through various stages \citep{xi2023rise}: from symbolic agents relying on symbolic logic \citep{ginsberg2012essentials, newell2007computer}; Reactive agents prioritizing environmental interaction and instantaneously responding \citep{brooks1991intelligence, brooks1986robust}; Reinforcement learning-based agents trained to handle complex tasks \citep{ribeiro2002reinforcement} but lacking generalization \citep{ghosh2021generalization}; Agents with transfer learning \citep{brys2015policy, zhu2023transfer} and meta-learning \citep{gupta2018meta, rakelly2019efficient} based on meta-learning and transfer learning to improve the generalization of the agent to the task. To the current LLM-based agents, where LLM is used as the brain of the agents \citep{park2023generative, sumers2024cognitive}. LLM can interpret inputs, plan output actions, and demonstrate reasoning even with the abilities of decision-making.

The emergence of ChatGPT \citep{dale2021gpt} has sparked a surge of interest in LLMs within the scientific research community and industry in recent years. LLMs possess exceptional capabilities, often serving as the brains of agents, and have zero-shot and few-shot generalization abilities that enable them to adapt to various tasks without parameter updates. Their natural language understanding and generation capabilities are unparalleled, allowing them to gain reasoning and planning abilities \citep{wei2022chain}. Additionally, LLMs can parse high-level abstract instructions to perform complex tasks without requiring step-by-step guidance\footnote{BabyAGI \url{https://github. com/yoheinakajima/babyagi}}, and their human-like text-generation capabilities make them highly effective communicators \citep{gravitas2023auto}. Furthermore, LLMs can sense their environment \citep{goodwin1995formalizing}, and technologies that expand their action space allow them to interact with the physical environment and complete tasks \citep{yin2023survey, zhu2023minigpt}. They also possess reasoning and planning capabilities, such as logical and mathematical reasoning \citep{wang2022self, wei2022chain}, task decomposition \citep{zhou2022least}, and planning \citep{xi2023self} for specific tasks. LLM-based agents have been used in various real-world scenarios \citep{li2023camel, qian2023communicative} and have shown potential for multi-agent interactions and social capabilities. Overall, LLMs have revolutionized the field of AI and hold great promise for future advancements.

\textbf{Capacity of LLM in Robotics.} LLM serves as the brain of the robot, functioning as the central component that integrates knowledge, memory, and reasoning capabilities to enable the robot to plan and execute tasks intelligently.

\textbf{Knowledge}. The knowledge of LLM for robotics can be categorized into two types: the knowledge that needs to be acquired through learning (which is the pre-trained dataset) and the knowledge that has been learned and stored in memory \citep{xi2023rise}.

\begin{itemize}
    \item \textbf{Pre-trained data.} There are various types of pretraining datasets available, and the more extensive and richer the knowledge learned, the stronger the LLM's generalization and natural language understanding capabilities will be \citep{roberts2020much}. Theoretically, the more a language model learns, the more parameters it possesses, enabling it to acquire complex knowledge in natural language and gain powerful capabilities \citep{kaplan2020scaling}. Research has shown that a richer dataset for language model learning can result in correct answers to a diverse range of questions \citep{roberts2020much}. Datasets can be categorized into different types, such as basic semantic knowledge, which provides an understanding of language meaning \citep{vulic2020probing}; Common sense, including everyday facts like people eating when hungry or the sun rising in the east \citep{safavi2021relational}; Professional field knowledge, which can aid humans in completing tasks like programming \citep{xu2022systematic} and mathematics \citep{cobbe2021training}.

    \item \textbf{Memory.} Just like human memory, embodied intelligence should be able to formulate strategies and make decisions for new tasks based on experiences (i.e., observed actions, thoughts, etc.). When faced with complex tasks, the memory mechanism can aid in reviewing past strategies to obtain more effective solutions \citep{hutter2000theory, squire1986mechanisms}. However, memory poses some challenges, such as the length of memory sequences and how to efficiently store and index them as the number of memories grows. As the robot's memory burden increases over time, it must be able to effectively manage and retrieve memories to avoid catastrophic forgetfulness \citep{kemker2018measuring}.
\end{itemize}

\textbf{Reasoning}. Reasoning serves as a foundational element in human cognition, playing a crucial role in problem-solving, decision-making, and the analytical examination of information \citep{wason1968reasoning, wason1972psychology}. Reasoning plays a crucial role in enabling LLMs to solve complex tasks. Reasoning capabilities allow LLMs to break down problems into smaller, manageable steps and solve them starting from the current status and known conditions. There is ongoing debate about how LLMs acquire their reasoning abilities, with some arguing that it is a result of pre-training or fine-tuning \citep{huang2022towards}, while others believe that it emerges only at a certain scale \citep{webb2023emergent}. Research has shown that Chain-of-Thought (CoT) \citep{wason1972psychology} can help LLMs reveal their reasoning capabilities, and some studies suggest that inference abilities may stem from the local static structure of the training data. SPINE \citep{ravichandran2024spine} has pioneered the development of a symbol-perception-action closed-loop system, enabling autonomous execution of complex tasks through a robust framework that leverages LLM-driven subtask inference, semantic composition, rolling planning, and validation. ELLMER \citep{mon2025embodied} revolutionizes robot capabilities by enabling the completion of complex, long-term tasks in uncertain and dynamic environments. By seamlessly integrating LLM with advanced sensor feedback mechanisms, ELLMER achieves efficient and accurate conversion of natural language instructions into actionable robot execution.

\textbf{Planning}. Humans engage in planning when faced with complex challenges. Planning can help people organize their thoughts, set goals, and determine appropriate actions in a given situation \citep{grafman2004planning, unterrainer2006planning}. Through planning, they can gradually work toward achieving their objectives. At its core, planning relies on reasoning. The agent can leverage reasoning capabilities to decompose high-level abstract instructions into executable subtasks and formulate rational plans for each subtask \citep{crosby2013automated, sebastia2006decomposition}. For example, LM-Nav uses ChatGPT to process received natural language instructions \citep{shah2023lm}. PaLM-E directly implements end-to-end processing, converting the received multi-modal input into multi-modal sentences for LLM processing \citep{driess2023palm}. Agents may also be able to reasonably update task planning based on the current situation through multiple rounds of dialogue and self-questioning, and answering in the future. Many studies have proposed methods of dividing the execution tasks into many executable small tasks during the planning process. For example, directly break down the execution task into many small tasks and execute them sequentially \citep{raman2022planning, xu2023rewoo}. CoT only processes one sub-task at a time and can adaptively complete the task, which has a certain degree of flexibility \citep{kojima2022large, wei2022chain}. There are also some vertical planning methods that divide tasks into tree diagrams \citep{hao2023reasoning, yao2023tree}. FLARE \citep{kim2025multi} empowers robots to generate environment-aware, executable plans with minimal language-based action examples, and to adaptively replan during execution. LLM+MAP \citep{chu2025llm+} pioneers a groundbreaking multi-agent planning mechanism for two-handed robots by integrating GPT-4o with PDDL, enabling the efficient and coordinated generation of plans from natural language descriptions to logically correct executions. 

\textbf{Generalist Robot Policy.} Large policies pretrained on diverse robot datasets have the potential to transform robotic learning: instead of training new policies from scratch, such generalist robot policies may be fine-tuned with only a small amount of in-domain data, yet generalize broadly. However, to be widely applicable across a range of robotic learning scenarios, environments, and tasks, such policies must handle diverse sensors and action spaces, accommodate a variety of commonly used robotic platforms, and fine-tune readily and efficiently to new domains. Octo \citep{team2024octo}, a large transformer-based policy trained on 800K trajectories from the Open X-Embodiment dataset, the largest robot manipulation dataset to date, lay the groundwork for developing open-source, widely applicable, generalist policies for robotic manipulation. It can be instructed via language commands or goal images and can be effectively fine-tuned to robot setups with new sensory inputs and action spaces within a few hours on standard consumer GPUs. 

\textbf{Low-level Trajectories Generation.} LLMs have recently shown promise as high-level planners for robots when given access to a selection of low-level skills. However, it is often assumed that LLMs do not possess sufficient knowledge to be used for the low-level trajectories themselves. Teyun Kwon et al. \citep{kwon2024language} address this assumption thoroughly and investigate if an LLM (GPT-4) can directly predict a dense sequence of end-effector poses for manipulation tasks, when given access to only object detection and segmentation vision models. They designed a single, task-agnostic prompt, without any in-context examples, motion primitives, or external trajectory optimizers. Their conclusions raise the assumed limit of LLMs for robotics, and they reveal for the first time that LLMs do indeed possess an understanding of low-level robot control sufficient for a range of common tasks. They can additionally detect failures and then re-plan trajectories accordingly.

\subsubsection{Multi-Robot Collaboration}

The significant advancements include language generation, understanding, and few-shot learning in LLMs have presented novel opportunities for tackling planning and decision-making within multi-agent systems. LLMs can enhance mathematical and strategic reasoning abilities through multi-agent debate \citep{du2023improving}. However, as the number of agents increases, the issues of hallucination in LLMs and coordination in multi-agent systems (MAS) have become increasingly pronounced \citep{zhang2023controlling}.

\textbf{Task Plan.} Recent work has demonstrated that pre-trained LLMs can be effective task planners. These studies primarily focus on single or multiple homogeneous robots on simple tasks. An under-explored but natural next direction is to investigate LLMs as multi-robot task planners. However, long-horizon, heterogeneous multi-robot planning introduces new challenges of coordination while also pushing up against the limits of context window length. It is therefore critical to develop token-efficient \citep{zhang2023controlling} LLM planning frameworks capable of reasoning about the complexities of multi-robot coordination. Chen et al. \citep{chen2024scalable} have compared the task success rate and token efficiency of four multi-agent communication frameworks (centralized, decentralized, and two hybrid) as applied to four coordination-dependent multi-agent 2D task scenarios for increasing numbers of agents. There are also some tasks with unique characteristics that can be a good task planner of multi-robot cooperation: (1) COHERENT \citep{liu2025coherent}, a novel LLM-based task planning framework for collaboration of heterogeneous multi-robot systems, including quadrotors, robotic dogs, and robotic arms. Specifically, a Proposal-Execution-Feedback-Adjustment (PEFA) mechanism is designed to decompose and assign actions for individual robots, where a centralized task assigner makes a task planning proposal to decompose the complex task into subtasks, and then assigns subtasks to robot executors. Each robot executor selects a feasible action to implement the assigned subtask and reports self-reflection feedback to the task assigner for plan adjustment. The PEFA loops until the task is completed. (2) Liu et al. \citep{liu2024leveraging} leverage the remarkable potential of LLMs to establish a decentralized heterogeneous ad hoc teamwork collaboration framework that focuses on generating reasonable policies for an ad hoc robot to collaborate with original heterogeneous teammates. A training-free hierarchical dynamic planner is developed using the LLM together with the newly proposed Interactive Reflection of Thoughts (IRoT) method for the ad hoc agent to adapt to different teams. (3) EMOS \citep{chen2024textbf}, a novel LLM-based MAS framework that first conducts embodiment-aware reasoning with a self-generated robot resume (Including textual and numerical descriptions of robot capabilities), rather than human-assigned role playing, to operate a collaborative HMRS. It includes a Centralized Group Discussion stage, where robots are assigned appropriate tasks based on their respective profiles.

\textbf{Task Execution.} Task execution is the next step of multi-robot collaboration. Roco \citep{mandi2024roco} is a novel approach for multi-robot collaboration that harnesses the power of pre-trained LLMs for both high-level communication and low-level path planning. Robots in Roco are equipped with LLMs to discuss and collectively reason task strategies. They generate sub-task plans and task space waypoint paths. EMOS \citep{chen2024textbf} demonstrated that decentralized Action Parallel Execution is more efficient. As a result of task allocation, each robot's dedicated agent begins to execute its actions in parallel. Given a subtask description and action execution history, the robot-specific agent controls the current agent through LLM FunctionalCall and the robot control library. These robot control libraries are implemented using ground truth world information, classic robot trajectory planners, and inverse kinematics solvers. When the intelligent agent completes all inference actions, it will automatically wait for the state. When the agent fails to complete a subtask but "believes" that it has completed it, it can continue to perform the following planned actions.

\subsection{Control}

Here, we argue that the control module is the key component responsible for regulating robotic actions. This module plays a crucial role in ensuring that the robot's actions are executed accurately and successfully, with a focus on the hardware aspects of action execution.

\subsubsection{Action Policy Based on Language}
To address the fundamental challenges in robot learning of natural language action policy, the research focuses on resolving two core issues:  How to Learn language-conditioned Behavior? and How to Execute Action After Parsing Natural Language?

\textbf{How to Learn Language-conditioned Behavior?} Much of the previous work has focused on enabling robots and other agents to comprehend and execute natural language instructions \citep{chen2011learning, duvallet2016inferring, luketina2019survey}. There are various approaches to learning linguistically conditioned behaviors, such as image-based behavioral cloning that follows the BC-Z \citep{jang2022bc} method or the MT-Opt \citep{kalashnikov2021mt} reinforcement learning method. Imitation learning techniques train policies on demonstration datasets \citep{jang2022bc, zhang2018deep}, while offline reinforcement learning has also been studied extensively \citep{jang2021gpt, kostrikov2021offline2, meng2021offline}. However, some works suggest that imitation learning on demonstration data performs better than offline reinforcement learning \citep{mandlekar2021matters}, and other studies demonstrate the feasibility of offline reinforcement learning in theory and practice \citep{kumar2022should, kumar2022pre}. Many works combine RL and Transformer structures \citep{chen2021decision,janner2021offline}, and some works integrate imitation learning with reward conditions, such as Decision Transformer (DT) \citep{chen2021decision}, namely combines imitation learning with reinforcement learning elements. However,  DT does not enable the model to learn from the demonstration dataset to have better performance. Deep Skill Graphs (DSG) \citep{bagaria2021skill} present a novel approach to skill learning utilizing the option framework. This method leverages graphs to represent discrete aspects of the environment, enabling the model to acquire structured knowledge and learn complex skills within the given domain.  CT employs goal-conditioned RL to transform the local skill-learning problem into a goal-conditioned Markov decision process (MDP) \citep{kaelbling1993learning}.

In the context of navigation robots, early approaches to enhancing navigation strategies with the natural language employed static machine translation \citep{lopez2008statistical} to discover patterns. The process involves utilizing discovery patterns to translate free-form instructions into formal languages that adhere to specific grammatical rules \citep{chen2011learning,matuszek2013learning, tellex2011understanding}. However, these methods were limited to structured state spaces. Recent works have also developed the VLN task as a sequence prediction problem \citep{anderson2018vision,mei2016listen,shimizu2009learning}. Additionally, there are methods that leverage nearly 1M labeled simulation trajectory demonstration data for training \citep{gu2022vision}, but applying these models in unstructured environments remains a significant challenge. Data-driven approaches for vision-based mobile robot navigation often depend on the utilization of realistic simulation techniques \citep{kolve2017ai2, savva2019habitat, xia2018gibson} or gathering supervised data to directly learn policies for achieving goals based on observations \citep{francis2020long}. Alternatively, self-supervised learning methods can utilize unlabeled datasets or trajectories generated automatically by onboard sensors and hindsight relabeling learning \citep{hirose2019deep, kahn2018self, shah2021ving}.

\textbf{How to Execute Action After Parsing Natural Language?} To determine whether a skill can be executed in the current state after parsing a natural language command, a temporal-difference-based (TD) reinforcement learning approach can be employed. This method learns a value function to evaluate whether the skill is executable or not \citep{ahn2022can}. The value function is derived from the corresponding affordance function of reinforcement learning \citep{gibson1977theory}. Additionally, LM-Nav \citep{shah2023lm} utilizes a self-supervised learning method to enhance the parsing of free-form language instructions by leveraging a pre-trained VLM across a large number of previous environments. To address the challenges of long-term tasks, hierarchical reinforcement learning (HRL) \citep{hutsebaut2022hierarchical} can be employed, where higher-level policies play a role in setting objectives for lower-level protocols to execute \citep{nachum2018data, vezhnevets2017feudal}. The process of mapping natural language and observations into robot actions can also be viewed as a sequence modeling problem \citep{brohan2023rt,brohan2022rt,rtx}. Transformer-based robot control, such as the Behavior Transformer \citep{shafiullah2022behavior}, focuses on learning demonstrations that correspond to each task. Gato \citep{reed2022generalist} suggests training a model on large datasets, including robotic and non-robotic.

\subsubsection{Vision-Language-Action Model}

\begin{figure*}[t]
    \centering
    \includegraphics[scale=0.50]{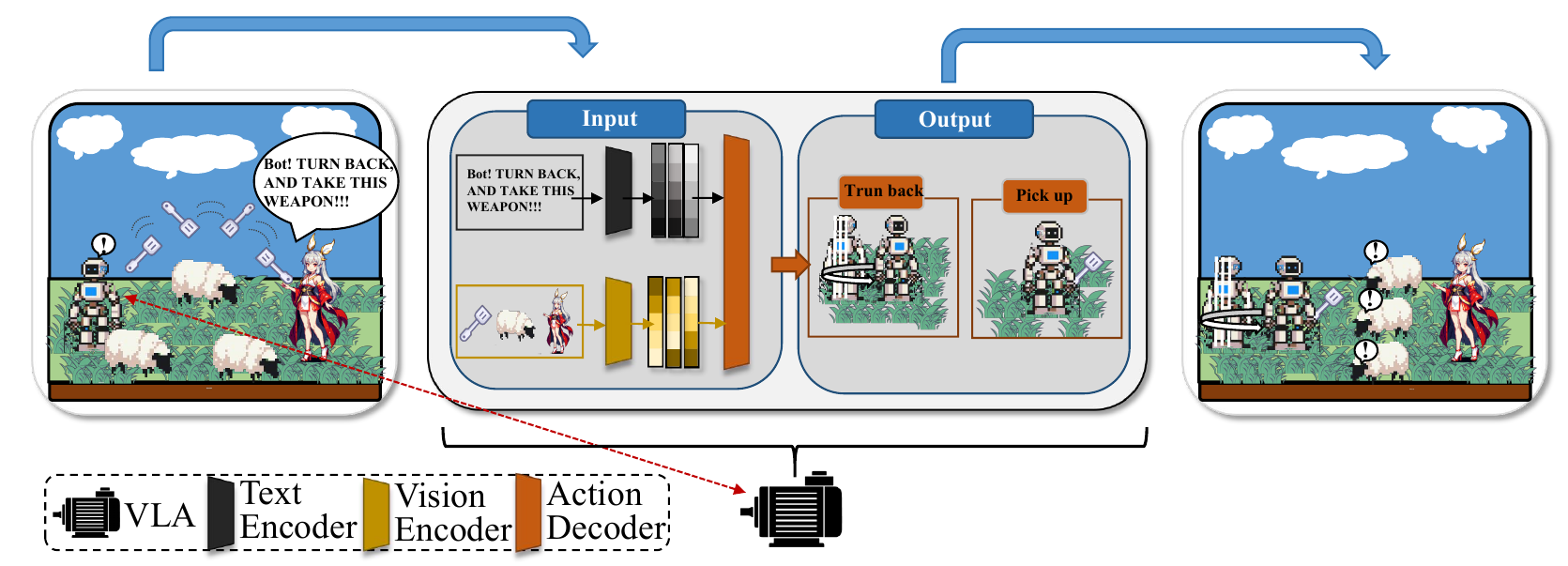}  
     \caption{The overview of the VLA model.}
    \label{fig: vla}
\end{figure*}

The VLA model derived from RT-1 \citep{brohan2022rt}, establishing an end-to-end model which integrates visual perception, language understanding, and action generation to directly output executable robotic commands. VLA enables robotics to perceive environmental stimuli,  interpret task requirements, and execute appropriate physical responses --- effectively bridging the gap between AI cognition and real-world interaction. We give a simple example to illustrate the structure of the VLA in Fig. \ref{fig: vla}. In the subsequent sections, we will present a systematic analysis of representative VLA developments in recent years, including OpenVLA \citep{kim24openvla}, \(\pi\) series \citep{black2024pi_0, pertsch2025fast, intelligence2025pi_}, Hirobot \citep{shi2025hi}, and Gr00t-N1\citep{bjorck2025gr00t}. We introduce these models in chronological sequence.

\textbf{OpenVLA} \citep{kim24openvla} is a VLA model developed through a collaboration between Stanford University, UC Berkeley, and the Toyota Research Institute (TRI). OpenVLA equips robotic systems with integrated multimodal capabilities encompassing visual perception,  natural language comprehension, and action generation, thereby enabling complex task execution through natural language instructions. OpenVLA employs a hybrid visual encoder combining DINOv2's \citep{OquabDMVSKFHMEA24} spatial feature extraction with SigLIP's \citep{ZhaiM0B23} semantic representation capabilities. These visual features are projected into linguistic space via an MLP while language processing utilizes LLaMA-2's \citep{touvron2023llama} pretrained representations. Notably, OpenVLA achieves performance parity with RT-2-X \citep{rtx} on Google's robotic test despite having an order-of-magnitude smaller parameter count.

\textbf{$\bm{\pi}$ series} represents a foundational general-purpose robotic model developed by the Physical Intelligence research team. (1) \(\pi_0\) \citep{black2024pi_0} is a flow model integrated generalist robot policy and VLA models.  \(\pi_0\) employs a standard Vision-Language Model (VLM) framework based on PaLI-Gemma \citep{paligemma24} to process visual and linguistic inputs, while robot-specific action and state data are handled by a dedicated Action Expert Module. This specialized design ensures the precise execution of high-frequency dexterous manipulation tasks. Specifically, \(\pi_0\)  employs conditional flow matching \citep{LipmanCBNL23} and action chunking \citep{ZhaoKLF23} for temporal abstraction. These components enable the model to generate continuous action sequences at 50Hz control frequencies. (2) \(\pi_0\)-FAST \citep{pertsch2025fast} is an optimized variant of \(\pi_0\). The paper proposes FAST, a novel tokenization method that leverages the Discrete Cosine Transform (DCT) for temporal action encoding. By integrating DCT with Byte Pair Encoding (BPE), Fast compresses original action sequences into compact, information-dense tokens. Benchmark results demonstrate a 10× improvement in token efficiency compared to conventional action tokenization methods. (3) $\pi_{0.5}$ \citep{intelligence2025pi_} significantly enhances the capabilities of VLA model, enabling it to excel in complex, long-term, and dexterous tasks within the dynamic and unstructured "open world" environment. $\pi_{0.5}$ is also the first VLA model to demonstrate the ability to complete intricate tasks in a completely unfamiliar home setting. $\pi_{0.5}$ leverages heterogeneous data co-training and adopts a hierarchical inference approach, enabling the generation of actions via tokens and action chunks. The training process is divided into two stages: pre-training and fine-tuning. During the pre-training phase, a mixed input of multiple robots and web data is utilized to train the transformer, generating sequence representations of text tokens, image patches, and action tokens. In the fine-tuning stage, the focus shifts to mobile manipulation tasks, incorporating action experts and Flow Matching. By leveraging a small amount of high-quality demonstrations.

\textbf{HiRobot} \citep{shi2025hi} has developed a general robot framework that integrates hierarchical VLA structures, enabling the robot to reason about complex instructions and feedback, as well as act on real-world physical movements. The HiRobot system consists of two primary components: a high-level policy and a low-level policy. The high-level policy is responsible for thinking complex instructions, decomposing them into next-step language instructions. The low-level policy focuses on executing specific actions, such as grasping a cup or placing it on a plate. During training, Hirobot divides the demonstration trajectory into short skill instructions through manual annotation and utilizes a large VLM to generate (contextual prompts, human feedback) data pairs, improving the high-level model's ability to handle complex instructions.  

\textbf{GR00T-N1} \citep{bjorck2025gr00t}, developed by NVIDIA, is the world's first open-source humanoid robot foundation model, significantly enhancing the capabilities of visual-language-action (VLA) models. It excels in complex, long-term, and dexterous tasks within dynamic, unstructured "open-world" environments and is the first VLA model to demonstrate the ability to complete intricate transactions in completely unfamiliar home settings. Leveraging heterogeneous data co-training and a hierarchical inference approach, GR00T-N1 enables the generation of auto-generated tokens and action chunks. The model adopts a unique dual-system architecture: ‌System 2 (Visual-Language Module)‌, based on the pre-trained Eagle-2 VLM, interprets environments and plans tasks, while ‌System 1 (Diffusion Transformer Module)‌ generates real-time, fluid motions with sub-300ms latency. These modules are tightly coupled through end-to-end joint training. The "data pyramid" used in the pre-training stage strategy—layering human behavior videos (unsupervised), synthetic data, and real robot data—ensures high data efficiency, requiring only 10\% labeled data to outperform traditional methods. We have shown the data pre-trained in GR00T-N1 in Appendix A. 

\subsubsection{Human Intervention}

When robots perform tasks, the model must generate specific action sequences, as the quality of these actions directly impacts the success rate and efficiency of task execution. The following introduces two approaches for generating action sequences: directly using VLA models and employing RL methods with human intervention. VLA can significantly enhance the generalization capabilities of robots. ECoT \citep{zawalskirobotic} pioneers the integration of chain-of-thought thinking reasoning into embodied intelligence, enabling robots to engage in deliberate and reflective decision-making before executing actions. By leveraging multi-step structured reasoning, ECoT significantly improves the quality and generalization capabilities of VLA. When using VLA models or RL to output robot actions, it often leads the robot to fall into local optimal strategies or irreversible states. SERL \citep{luo2024precise} is a method for implementing precise and dexterous robotic manipulation via human-in-the-loop reinforcement learning, which integrates demonstration and human intervention, efficient RL algorithms, and other system-level design choices, achieving near-perfect success rates and fast cycle times within only 1 to 2.5 hours of training time.

\subsection{Interaction}

Interaction serves as a fundamental module that enables robots to engage and interact with both the environment and humans. To enhance robots' ability to interact in the physical world, they are often trained extensively. While some researchers utilize AI to interact in virtual environments, such as games or simulations, ultimately, these models must be transferred to the real world. However, the accuracy of these models tends to be lower in real-world settings compared to simulated environments.

\subsubsection{Game}

Recent studies have focused on constructing generalist agents that can operate effectively in open-world environments (such as Minecraft \citep{arkin2020multimodal}). Most of these studies focus on material collection and tool-crafting, with the Obtain Diamond task often being treated as the ultimate objective. After integrating LLMs or MLLMs as a brain for planning and reasoning, these agents become more efficient in achieving their final objectives. 

Most recent studies have used LLMs as a cognitive component to plan tasks in Minecraft. VOYAGE \citep{wang2023voyager}, a pioneering LLM-powered embodied agent in Minecraft, enables continuous exploration, acquisition of diverse skills, and novel discoveries, all without human intervention. It leverages the vast knowledge base of GPT-4 by prompting it to generate a continuous stream of novel tasks and challenges, which are then used to construct a Skill Library that serves as the foundation for learning and evolution. This process enables self-improvement through a feedback loop, where the agent refines its performance based on environmental feedback, execution errors, and self-verification, ultimately driving program improvement. OBDYSSEY \citep{liu2024odyssey} fine-tunes a LLaMA-3 model to create MineMA, a specialized model trained on a large question-answering dataset comprising over 390,000 instruction entries sourced from the Minecraft Wiki, and subsequently utilizes MineMA as an LLM planner. GITM \citep{zhu2023ghost} presents a novel architecture comprising the LLM Decomposer, LLM Planner, and LLM Interface, which collaboratively decompose high-level goals into sub-goals, structured actions, and actionable keyboard/mouse operations. This approach enables GITM to successfully acquire all items in the game. DECKARD \citep{nottingham2023embodied} conceptualizes the LLM as a tabula rasa world model, treating it as the hypothesized Abstract World Model (AWM), and then optimizes it in Minecraft by leveraging a directed acyclic graph (DAG) to record and update skills in a structured and hierarchical manner.

Some studies employ MLLMs as a cognitive component to achieve the ultimate objective. JARVIS-1 \citep{wang2025jarvis} presents a novel multi-task agent, specifically designed to navigate the complex environment of Minecraft. Notably, JARVIS-1 effectively interprets multimodal inputs and adeptly translates them into actionable decisions. JARVIS-1 also features a multimodal memory component, which enables it to retain observations and environmental feedback. DEPS \citep{wang2023describe} enhances error correction in initial LLM-generated plans by incorporating a descriptive analysis of the plan execution process and providing self-explanatory feedback upon encountering failures during extended planning phases. Furthermore, it employs a state-aware selector based on VLM to dynamically select the most suitable goal given the current state.

\subsubsection{Language-based Human-robot Interaction}

There are GUI (Graphical User Interface) and LUI (Language User Interface) for human-robot interaction. GUI refers to a computer-operated user interface that is graphically displayed and uses an interactive device to manage the interaction with the system. Unlike GUI, LUI can directly use natural human language for human-robot interaction, and the most representative LUI product is ChatGPT. Traditionally, the task of simulating human-robot interaction using natural language has proven to be difficult due to the constraints imposed on users by rigid instructions or the need for intricate algorithms to manage numerous probability distributions related to actions and target objects \citep{arkin2020multimodal}. However, it is not easy to translate instructions into commands that robots can understand in the real world, and traditionally, fixed collections of desired actions and directives have been used to enable robots to understand human language. However, this can significantly limit the robot's flexibility and has limited generalizability across different hardware platforms. The Language Trajectory TransformEr \citep{bucker2023latte} introduces a versatile, language-driven framework that empowers users to customize and adapt the overall trajectories of robots. The approach leverages pre-trained language models (e.g., BERT \citep{devlin2018bert} and CLIP \citep{radford2021learning}) to encode the user's intentions and target objects directly from unrestricted text inputs and scene images. It combines geometric features produced by a network of transformer encoders and generates the trajectories using a transformer decoder, eliminating the need for prior task-related or robot-specific information.

Considering the vagueness and ambiguity of natural language, from the point of view of human-robot interaction, robots should enhance the initiative of interaction in the future; that is to say, let the robot actively ask the user questions through the LLMs. If the robot feels that the user's words are problematic and is not sure what they mean, it should ask you back what you mean or whether you mean what you say.

\section{Cross-Module Coordination} 
\label{sec: cross-module coordination}

In this section, we focusing on how the module introduce in Section \ref{sec:technologies} interacts and synergizes to address real-world challenges. We delve into the intricacies of their collaborative functionality, highlighting how the integration of these modules enables the effective solution of practical problems.

\subsection{Affordance Grounding} 

Affordance grounding typically involves associating objects in the environment with their possible uses or ways of interaction, necessitating the integration of perception and reasoning capabilities. Through this association, models can more effectively understand and predict how to interact with objects in the environment, thereby enhancing their autonomy and adaptability. It forms the stepping stone to downstream tasks such as understanding human-object interaction, visual navigation, and object manipulation. Affordance grounding is a fundamental but challenging task, as a successful solution requires the comprehensive understanding of a scene in multiple aspects, including detection, localization, and recognition of objects with their parts; of the geospatial configuration/layout of the scene; of 3D shapes and physics; as well as of the functionality and potential interaction of the objects and humans. Much of the knowledge is hidden and beyond the image content, with the supervised labels from a limited training set. AffordanceLLM
\citep{qian2024affordancellm} attempt to improve the generalization capability of the current affordance grounding by taking the advantage of the rich world, abstract, and human-object-interaction knowledge from pretrained large-scale vision language models. Visual Affordance-guided Policy Optimization (VAPO)
\citep{AffordanceLearning2} extracts a self-supervised visual affordance model from human teleoperated play data and leverages it to enable efficient policy learning and motion planning, which combines model-based planning with model-free deep RL to learn policies that favor the same object regions favored by people, while requiring minimal robot interactions with the environment.  GREAT \citep{shao2025great} introduces a pioneering collaborative reasoning framework that seamlessly integrates the geometric properties of objects with interaction intentions, thereby enriching affordance knowledge through analogical reasoning.

\subsection{Physical Grounding}

Physical grounding involves the precise correspondence of object or language information, as perceived by robots, to specific entities or physical states in the real world. This process requires the integration of perception and spatial reasoning capabilities. VLMs are particularly well-positioned to reason about the physical world, especially within domains such as robotic manipulation. However, current VLMs are limited in their understanding of the physical concepts (e.g., material, fragility) of common objects, which restricts their usefulness for robotic manipulation tasks that involve interaction and physical reasoning about such objects. PHYSOBJECTS \citep{gao2024physically}, an object-centric dataset of 39.6K crowd-sourced and 417K automated physical concept annotations of common household objects, can help robots to understand the physical concepts of the manipulation targets. VLMs fine-tuning on PhysObjects improves their understanding of physical object concepts, including generalization to held-out concepts, by capturing human priors of these concepts from visual appearance. Physically grounded VLMs integrated into an interactive framework with an LLM-based robotic planner demonstrate improved planning performance on tasks requiring reasoning about physical object concepts. Additionally, these models achieve higher task success rates in real robots equipped with physically grounded VLMs. GLIMO \citep{liu2024grounding} leverages simulators and LLMs to generate diverse, interactive data, enabling the training of LLMs to acquire common sense and causal understanding of the physical world. Notably, GLIMO's self-reinforcement learning mechanism enhances the physical reasoning and task execution capabilities of LLMs, even in scenarios where world models are incomplete or inaccurate. 

\subsection{Detect and Analyze Failed Executions}

The ability to detect and analyze failed executions automatically is crucial for an explainable and robust robotic system. Additionally, the ability necessitates the integration of modules that combine perception, action monitoring, and causal reasoning capabilities. REFLECT \citep{liu2023reflect}, a framework that queries LLM for failure reasoning based on a hierarchical summary of robot's past experiences generated from multisensory observations, leverages the strong reasoning abilities on textual inputs of LLMs for robot failure explanation, which can further guide a language-based planner to correct the failure and complete the task. AHA \citep{duan2025aha} fault recognition system is grounded in the VLM, enabling the detection of failures and the generation of explanatory insights. By leveraging the Failed Gen approach, researchers have automatically generated a large-scale dataset of failed trajectories, which is used to train AHA to identify errors and provide explanations for their causes.

\subsection{Navigation}

Navigation tasks require the use of both the robot’s perception and policy generation capabilities. For navigation tasks, a common solution is to add an additional navigation modality for large models, which directly learn navigation implications.

\textbf{Vision-navigation model.} The Berkeley Autonomous Driving Ground Robot (BADGR) \citep{kahn2021badgr} is a mobile robot navigation system that leverages end-to-end learning and self-supervised, non-policy data collected in real-world environments to train its algorithms without any simulation or human supervision. This innovative approach enables BADGR to navigate complex environments efficiently, paving the way for future advancements in autonomous driving technology. ViNG \citep{shah2021ving} is a goal-condition model inspired by GoalConditionedRL \citep{eysenbach2019search}. It is capable of predicting the temporal distance between image pairs and the corresponding actions to be performed. By integrating learned policies with topological maps constructed from previously observed data, ViNG's system can effectively determine how to achieve visually indicated goals, even in the presence of variable appearance and lighting conditions. RECON \citep{shah2021rapid} is a system for robot learning designed for exploring autonomously and navigating in complex and unpredictable real-world surroundings. The core of RECON leverages a latent variable model of learning distance and action, along with non-parametric topology memory, to enable efficient and effective exploration. ViKiNG \citep{shah2022viking}, built upon RECON mapping, incorporates geographical hints to propose an integrated learning and planning method that utilizes auxiliary information. This method combines a local traversability model. The model evaluates the robot's present camera observation and utilizes a potential sub-goal to infer the difficulty of achieving it. With a heuristic model that examines hints in the cost graph and evaluates the suitability of these sub-goals in achieving the overall goals, the general navigation model (GNM) \citep{shah2023gnm} aims to train a general goal-condition model for vision-based navigation that can broadly generalize across diverse environments and embodiments, leveraging data from multiple structurally similar robots. By developing pre-trained navigation models with such capabilities, GNM represents a significant step toward realizing this vision that envisions applications for new types of robots.

\textbf{Vision-and-language navigation model.} One of the primary objectives of AI research is to develop an embodied intelligence that can effectively communicate with humans and interact with the environment. This embodied intelligence is capable of understanding human language and navigating its surroundings with ease, which has the potential to greatly benefit human society. However, achieving this goal is not without its challenges, including insufficient datasets, navigation processing strategies, processing of multi-modal inputs, and model migration from familiar environments to unfamiliar environments. Despite these obstacles, the development of embodied intelligence remains a crucial area of research in the field of AI \citep{gu2022vision}. Visual-and-language navigation (VLN) is a model that leverages visual observations to directly learn navigation implications and seamlessly links images and actions across time. As an extension of visual navigation in both real environments \citep{mirowski2018learning} and simulated \citep{zhu2017target}, VLN boasts the capability to navigate complex 3D environments. There are many datasets in VLN that can be exploited. DyNaVLM \citep{ji2025dynavlm} is an end-to-end visual language navigation system that enables dynamic selection of navigation target perspectives, offering a flexible and adaptive approach that diverges from traditional fixed-step strategies. BrainNav \citep{ling2025endowing} introduces a groundbreaking navigation system that replicates the human-like cognitive structure, mirroring the neural mechanisms that govern spatial understanding and behavioral planning in the human brain. 

\section{Applications of LLMs in Robotics} \label{sec: applications}

Applications of large models and robotics across various domains. Here are ten specific applications of the combination of large models and robotics, along with their explanations:

\begin{itemize}
    \item \textbf{Autonomous navigation and path planning.} Large models provide powerful semantic understanding and reasoning capabilities for robots, assisting them in autonomous navigation and path-planning in unknown environments. By combining large models with sensor data, robots can comprehend semantic information in the environment, recognize obstacles, target locations, and navigation objectives, and generate suitable path-planning solutions \citep{crespo2020semantic}.

    \item \textbf{Speech interaction and NLP.} LLMs excel in speech recognition, semantic understanding, and natural language generation. Robots can leverage large models for speech interaction, understanding, and answering user queries, executing specific tasks, and providing personalized service experiences \citep{reed2022generalist}.

    \item \textbf{Visual perception and object recognition.} Large models possess strong capabilities in image and video analysis, aiding robots in object recognition, target detection, and scene understanding. By integrating deep learning and large models, robots can achieve efficient and accurate visual perception, which can be applied in autonomous driving, robot vision-based navigation, and industrial automation \citep{hu2022scaling, zhou2025physvlm}.

    \item \textbf{Human-robot collaboration and social robots.} Large models with natural language processing and emotion analysis help robots understand human feelings and intentions better, making interactions between humans and robots more natural and smart. Social robots can engage in conversations, comprehend emotions, and provide companionship and support, which are applied in fields like healthcare, education, and entertainment \citep{park2023generative}.

    \item \textbf{Humanoid robots and emotional expression.} Large models can help humanoid robots better understand and express emotions. Through natural language generation and emotion recognition technologies, robots can engage in emotional communication and expression with humans, providing emotional support and companionship \citep{liu2024unlocking}.

    \item \textbf{Industrial automation and robot control.} Large models can be combined with sensor data for industrial process monitoring, anomaly detection, and predictive maintenance. By learning and analyzing large-scale data, robots can achieve intelligent industrial automation and adaptive control \citep{brohan2022rt, brohan2023rt}.

    \item \textbf{Healthcare and rehabilitation robots.} Large models can be applied in medical and rehabilitation robots to assist with diagnosis, treatment, and patient care. Robots can analyze medical images, patient data, and clinical records, aiding in disease detection, surgical planning, and personalized therapy. They can also provide physical assistance and rehabilitation exercises for mobility-impaired patients \citep{pathak2025soft}.

    \item \textbf{Environmental monitoring and exploration.} Large models can be combined with robotic platforms for monitoring and exploration in various environments, such as oceans, forests, and disaster sites. These robots can analyze sensor data, satellite imagery, and other environmental information to monitor pollution levels, detect natural disasters, and explore uncharted territories \citep{lin2025bip3d}.

    \item \textbf{Agriculture and farm mechanization.} Large models and robots can be applied in agriculture and farm mechanization, optimizing crop management, monitoring plant health, and automating labor-intensive tasks. Robots equipped with sensors and cameras can collect data from farmlands and analyze soil conditions, climate changes, and crop requirements, providing farmers with decision support to enhance agricultural productivity and sustainability \citep{zhao2023survey}.

    \item \textbf{Education and learning assistance.} Large models and robots can provide personalized tutoring and learning support in the field of education. Robots can interact with students, and then offer personalized learning materials and guidance based on their abilities and needs \citep{alam2022social}. Leveraging the semantic understanding and knowledge reasoning capabilities of large models, robots can answer questions, explain concepts, and help students deepen their understanding of knowledge.
\end{itemize}

In summary, the combination of large models and robotics holds tremendous potential across various domains, including autonomous navigation, speech interaction, visual perception, human-robot collaboration, industrial automation, healthcare, environmental monitoring, agriculture, and education. This integration can bring significant convenience and innovation to human life and work.

\section{Challenges} \label{sec: challenge}
\subsection{Datasets}

In the realm of Web 3.0 \cite{gan2023web}, big data \cite{sun2022big}, AI-Generated Content (AIGC) \cite{wu2023ai}, and machine learning, collecting datasets has always been a challenge. Currently, training LLMs requires vast amounts of data to support their capabilities, particularly high-quality datasets that consume considerable resources. In the field of robotics, collecting datasets is even more difficult. While LLM like ChatGPT relies on text data for pre-training \cite{brown2020language}, VLM uses a combination of text and image data \cite{radford2021learning}. Robotics, however, requires a combination of both, with the addition of multimodal data, such as text, images, and touch, to serve as the robot's sensory input. These diverse datasets need to be processed in a unified format \cite{driess2023palm}, allowing the robot's brain to plan and divide tasks effectively. Unfortunately, there is a lack of ready-made, multi-modal datasets, and collecting them requires a significant time investment. Moreover, policy control is necessary, which includes the interaction between the robot and its environment, necessitating 3D data \cite{overviewofagents}. The data required for robotics are diverse and scarce, with poor general applicability. For instance, a dataset used to train robot dogs cannot be applied to humanoid robots, and a dataset used for screwing in an assembly line may not be suitable for robots that assemble items. However, with the emergence of platforms similar to Open X-embodiment\footnote{Open X-embodiment repository, a dataset consisting of different platforms. \url{https://robotics-transformer-x.github.io/}}, the challenges of dataset collection in robotics may be alleviated in the future.

\subsection{Training Scemes}

As embodied intelligence necessitates interaction with the physical environment, the model's training requires specific scenarios, e.g., distributed training \cite{zeng2023distributed}. Current research involves training robot-related models in various environments, such as games \cite{park2023generative}, simulations \cite{devin2017learning}, and real-world scenarios \cite{bharadhwaj2023roboagent}. Training in-game scenarios is straightforward, with simple operations like button-pressing. However, the knowledge gained from games may not translate well to real-world scenarios, as the information in complex scenes varies greatly, and language models cannot provide a universal solution. Simulation environments aim to closely replicate reality, with low energy consumption and cost. However, modeling real scenes in simulators can be necessary. While game and simulation environments can train models, they share a common issue: poor transferability to real scenes. For instance, a model with 90\% accuracy in a game or simulation may only have 10\% accuracy in a real scene. Real-scene training faces significant challenges, such as cost. In simulations, objects can be generated through code \cite{chen2023interact}, but in reality, purchasing them can be expensive. Transferring models between different training scenarios is a significant challenge.

\subsection{Shape}
\begin{figure*}
    \centering
    \subfigure[Robotics with different shapes.]{
        \label{fig:shape}
        \includegraphics[scale = 0.3]{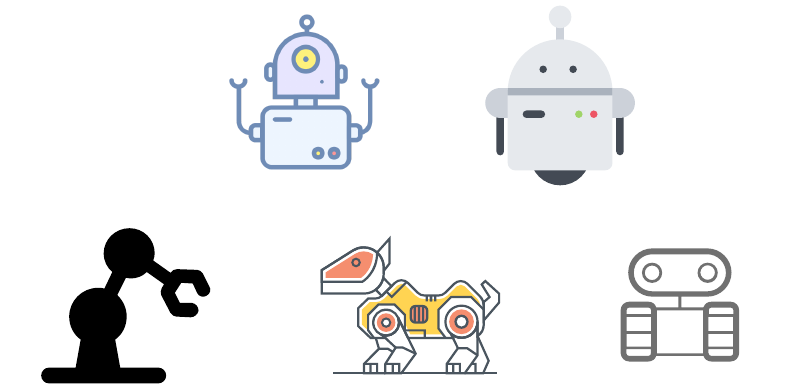}
    }
    \subfigure[LLM development.]{
        \label{fig: development}
        \includegraphics[scale = 0.7]{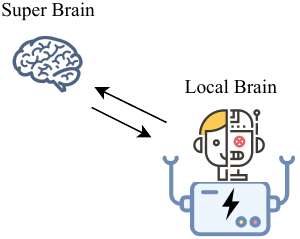}
    }
    \subfigure[Modualize.]{
        \label{fig: modualize}
        \includegraphics[scale=0.4]{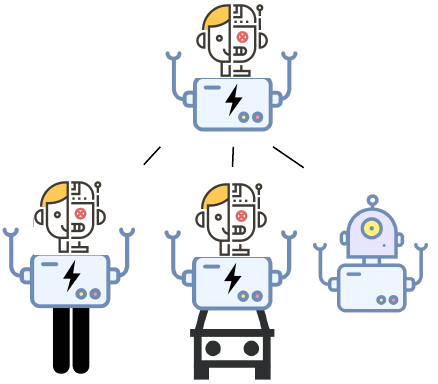}
    }
    
    \caption{Challenge in embodied intelligence.}
    \label{fig: Challenge in embodied intelligence}
\end{figure*}

Currently, most work environments in human society are well-suited for humanoid robots. However, the question arises: must robots be human-shaped \cite{hwang2013effects}. There are numerous types of robots in existence, each with unique capabilities and applications, as illustrated in Figure \ref{fig:shape}. From an energy consumption perspective, wheels are more energy-efficient than legs. Therefore, if a humanoid robot is built, using legs to move objects instead of a conveyor belt may be inefficient. Similarly, a chef robot may not need to hold a shovel and cook like a human. In many cases, designing a pipeline tailored to the specific task at hand can achieve more efficient automation than relying on humanoid robots. While humanoid robots are often depicted in animation scenes, such as in animations like Mobile Suit Gundam or games like Armored Core, their design may not always be practical for applications. For instance, a robot designed solely for washing dishes may not need the ability to sing. Modular concepts like Expedition A1\footnote{\url{https://www.agibot.com}}, can offer optimal results for different scenarios by replacing certain components. The shape of the robot remains a topic of debate, and the decision should ultimately focus on suitability for the task at hand.

\subsection{LLM Deployment}

Given embodied intelligence, the question arises regarding the deployment of its brain. Current technical limitations prevent the LLM from being deployed locally on the robot. The prevailing industry practice involves employing two brains: a cloud-based super brain and a local brain, like in Figure \ref{fig: development}. However, a unified consensus on this device-side plus cloud testing deployment method has yet to be established. A feasible solution could be to create a dynamic, compact model on the local client side, capable of handling basic scenario interactions. The cloud-based super brain, on the other hand, would tackle complex and challenging problems. The LLM deployment architecture remains a pressing issue that must be addressed in the future development of agents. This deployment structure also introduces latency issues, as information exchange between the robot and the super brain requires signal transmission. In certain environments, such as those with signal loss, the robot may be left with only its local brain, potentially leading to control loss or unpredictable behavior.

\subsection{Security}

LLMs like ChatGPT may harbor biases or misconceptions stemming from their pre-training data. These biases can manifest in problematic guidance for users, and robots that rely on LLMs as their brains may also exhibit biases \cite{xi2023rise}. Since robots' outputs are typically physical actions, biased or misunderstood guidance can lead to harmful consequences for users \cite{elhage2021mathematical,qin2023tool}, such as a chef robot burning down a house while cooking. Beyond physical safety risks, robots also raise concerns about data security \cite{mccallum2023chatgpt}. For instance, a robot butler who resides in a home may become intimately familiar with the household's environment and occasionally require cloud interaction for certain tasks. During user interaction, there is a risk of private data leakage, which could be mitigated by an offline environment, but this may compromise the robot's performance.

\subsection{Dialogue Consistency}

Humans often don't complete tasks in a single, static step. Instead, they iteratively adjust strategies and goals based on feedback received after taking action. The same is true for embodied intelligence. When faced with high-level, abstract, or ambiguous commands, robots may not be able to decompose them into executable small tasks at first. They need to obtain further feedback from the environment and humans through continuous dialogue to update their goals. Without this ability to engage in continuous dialogue, which enables robots to perform tasks dynamically, their performance is significantly impaired \cite{song2020generating}. Moreover, the maximum length limit of a robot's context is another issue worth considering. Typically, embodied intelligence may serve as a housekeeper, handling daily tasks such as washing dishes or drying clothes. However, for long-term tasks like scientific research, robots require enhanced context-understanding capabilities. Currently, there's a limit to the length of context that robots can handle, and this limitation can lead to catastrophic forgetting \cite{kemker2018measuring}. Dialogue persistence is a crucial challenge for long-term tasks.

\subsection{Social Influence}

The rapid advancement of LLMs is bringing the era of embodied intelligence, as depicted in science fiction movies and games, closer to reality. This technological breakthrough will undoubtedly revolutionize human society and unleash unprecedented productivity. With robots capable of performing repetitive tasks, the need for human labor in various industries will diminish. However, this shift may also have far-reaching consequences, potentially disrupting social structures and stability \cite{helberger2023chatgpt}. As robots replace low-end manual labor, it raises questions about the fate of those who previously held these jobs. The double-edged sword of embodied intelligence presents both liberation and disruption. While automation may usher in unprecedented efficiency, it also poses challenges for societal adaptation. Some works of science fiction, such as Detroit Become Human \footnote{\url{https://store.steampowered.com/app/1222140/}}, depict a future where robots gain consciousness and conflict with humans, leading to a war between the two. Alternatively, technology may fall into the wrong hands, becoming a tool for exploitation and solidifying class divisions. However, in a worst-case scenario, robots may become a replacement for humans. As we embrace the development of embodied intelligence, we must also confront the ethical and societal implications it entails.

\subsection{Ethic}

Embodied intelligence has long been regarded as a mere tool, but it may hold greater significance for some users. For instance, companion robots can provide solace to lonely individuals, much like a loyal friend. In fact, some people even develop emotional attachments to their first car or a vehicle that has been with them for a long time. If we were to create robots that resemble humans or exhibit human-like intelligence, would they evoke different emotions? In science fiction movies, robots that gain self-awareness and break free from their programming often develop emotions and even form relationships with humans. Interestingly, robots powered by LLMs have already demonstrated a degree of intelligence. Will they eventually become conscious? If embodied intelligence evolves to possess consciousness, should we still consider them tools? This raises questions about the definition of conscious robots and whether they can be considered human. Although this challenge is still far off in the future of smart robot development, it is an intriguing topic to ponder.

\section{Promising Directions for Future Work} \label{sec:directions}

\subsection{Security of Task Executing}

Security has always been a pressing concern in various models, particularly regarding user privacy. However, we argue that the safety of agents during task execution is of paramount importance. In this article, we explore whether an agent's actions during task execution could cause harm \citep{qin2023tool, elhage2021mathematical}. For instance, consider a scenario in which a robot is asked to make lunch, but in the process, sets the kitchen on fire. In other scenes, imagine a robot tasked with killing fish, but it mistakenly identifies humans as fish and proceeds to chase and harm them. These scenarios highlight the need to limit the actions an agent can perform to prevent potential harm. Current robot systems focus on enabling the robot to determine which actions can be performed based on the current state and environment, without fully considering the consequences of executing those actions. Therefore, we propose that ensuring the safety of task execution must be a top priority by guaranteeing that the robot's actions do not harm human rights and interests.

\subsection{Training Scenario Transfer}

Due to technical or economic constraints, it is common to train robot action policies in simulated \citep{devin2017learning} or gaming environments \citep{park2023generative}. However, the ultimate goal of agent training is to apply it in real-world scenarios. Unfortunately, training in diverse scenarios can lead to not being acclimatized, which may compromise the agent's performance when deployed in real-world situations. The fundamental source of this problem can be attributed to the disparity of feedback mechanisms between simulated and real-world environments. In games or simulations, feedback is often more straightforward, with the robot receiving clear and concise information about the outcome of its actions. In contrast, real-world feedback is more complex and nuanced, making it challenging to assess the feasibility of a task in a limited scenario. Therefore, a valuable research direction is to explore methods for transferring model training across different scenarios while maintaining their accuracy in the original training environments.

\subsection{Unify Format of Modal}

Currently, many models are utilizing LLM as the robot's brain, and text-type data is typically the input that LLM accepts. However, for agents reliant on multi-modal perception, efficiently handling diverse input formats poses a significant challenge. To address this issue, a VLA model has been proposed \citep{brohan2023rt}, which uniformly converts visual and natural language multi-modal inputs into multi-modal sentences for processing, and outputs actions in the same format. In other words, multi-modal statements are employed to harmonize input and output. Nevertheless, there is currently no unified processing for other modalities such as touch and smell. It is anticipated that unified multi-modal models like VLA will gain popularity in the future.

\subsection{Modular Components}

As previously discussed, the field of robotics currently lacks a unified approach to robot design, with varying opinions on the matter. We believe that there should be a modular design methodology, in which each component of the robot can be swapped out like a machine, just like in Fig. \ref{fig: modualize}, allowing for greater versatility and adaptability\footnote{\url{https://www.agibot.com}}. To achieve this, it is essential to first establish unified specifications for the various robot modules. For instance, a robot can be composed of a head, torso, upper limbs, and lower limbs, with the upper limbs and lower limbs being interchangeable based on the task at hand. Among them, the upper limbs and lower limbs can be replaced according to specific tasks. When we need to cook, we can use our upper limbs as a shovel, and when we need to deal with weeds in the yard, we can use our lower limbs as a weeder.

\subsection{Autonomous Perception}

Our current research focuses on developing robots that can interact with humans using natural language instructions. In many cases, we study how humans issue instructions and how robots can decompose abstract tasks into specific sub-tasks for execution \citep{ahn2022can}. However, we also hope that robots can perceive and respond autonomously to handle our current needs. For instance, if our cup falls to the ground and breaks, an agent should be able to perceive the situation through hearing and vision, and then autonomously handle the glass fragments for us. Autonomous perception requires the robot to have common sense, which is a capability that can be integrated into robots based on LLM as the brain. Research on robots' autonomous perception capabilities is crucial for improving our quality of life in the future.

\section{Conclusion}\label{sec:conclusion}

In this survey, we summarized the methods and technologies currently used for robots based on LLM. First, we review the development of LLM and explain what improvements will be brought to robots by using LLM as a brain. We also introduce the representative LLM-based robot models that have been proposed in recent years. Next, we divide the robot into four modules: perception, decision-making, control, and interaction. For each module, we discuss the relevant technologies and their functions, including the perception module's ability to process the robot's input from the surroundings; the decision-making module's capacity to understand human instructions and plan; the control module's role in processing output actions; and the interaction module's ability to interact with the environment. We also explore the potential application scenarios of current robots based on LLMs and discuss the challenges, such as training, safety, shape, deployment, and long-term task performance. Finally, we consider the social and ethical implications of post-intelligent robots and their potential impact on human society.

As LLMs continue to evolve, robots may become increasingly intelligent and capable of processing instructions and tasks more efficiently. With advancements in hardware, robots could eventually become reliable assistants for humans, as depicted in science fiction movies. However, we must also be mindful of their potential societal impact and address any concerns proactively. Embodied intelligence represents a new paradigm for the development of intelligent systems, holding significant implications for shaping the future. LLM-based robotics offers a potential path toward embodied intelligence. We hope this survey provides inspiration to the community and facilitates research in related fields.

\section*{Acknowledgment}
This research was supported in part by the National Natural Science Foundation of China (No. 62272196), the Guangzhou Basic and Applied Basic Research Foundation (No. 2024A04J9971), and the Natural Science Foundation of Guangdong Province (No. 2022A1515011861).

\section*{Declarations}
The authors declare that they have no known competing financial interests or personal relationships that could have appeared to influence the work reported in this paper.

\bibliographystyle{plainnat}
\bibliography{LLM4Robotics.bib}

\end{document}